\documentclass{article}

\usepackage[nonatbib,final]{neurips_2022}
\usepackage[numbers]{natbib}

\usepackage[utf8]{inputenc} 
\usepackage[T1]{fontenc}    
\usepackage{hyperref}       
\usepackage{url}            
\usepackage{booktabs}       
\usepackage{amsfonts}       
\usepackage{nicefrac}       
\usepackage{microtype}      
\usepackage{xcolor}         

\usepackage{microtype}
\usepackage{graphicx}
\usepackage{subfigure}
\usepackage{booktabs} 

\usepackage{hyperref}

\usepackage{hhline}
\usepackage{multirow}

\usepackage{amsmath}
\usepackage{amssymb}
\usepackage{mathtools}
\usepackage{amsthm}

\usepackage[capitalize,noabbrev]{cleveref}

\usepackage{caption}

\theoremstyle{plain}
\newtheorem{theorem}{Theorem}[section]

\theoremstyle{definition}
\newtheorem{definition}[theorem]{Definition}

\theoremstyle{remark}

\usepackage[textsize=tiny]{todonotes}

\usepackage{float}

\newcommand{\fat}[1]{\ifmmode\bm{#1}\else\textbf{#1}\fi}

\newcommand{\order}[1]{\mathcal{O}\left( #1 \right)}

\newcommand{\set}[1]{\mathbb{#1}}

\newcommand{\vect}[1]{\fat{#1}}

\newcommand{\matr}[1]{\fat{#1}}

\newcommand{\tens}[1]{\mathcal{#1}}

\newcommand{\func}[1]{\textsf{#1}}

\usepackage{mathtools}
\usepackage{url}
\usepackage{bm}
\usepackage{subfiles} 

\newcommand{\vx}[0]{\vect{x}}            
\newcommand{\vy}[0]{\vect{y}}            
\newcommand{\vz}[0]{\vect{z}}            
\newcommand{\vt}[0]{\vect{\theta}}       

\newcommand{\mw}[0]{\matr{W}}            
\newcommand{\mx}[0]{\matr{X}}            
\newcommand{\mz}[0]{\matr{Z}}            

\newcommand{\tj}[0]{\tens{J}}            

\newcommand{\fj}[0]{\func{J}}            

\usepackage[ruled,vlined,algosection,linesnumbered]{algorithm2e}

\title{TTOpt: A Maximum Volume Quantized Tensor Train-based Optimization and its Application to Reinforcement Learning}

\author{%
    Konstantin~Sozykin 
    \thanks{Equal Contribution, corresponding emails \{konstantin.sozykin,a.chertkov\}@skoltech.ru }
    \hspace{1pt}
    \thanks{Center of Artificial Intelligence  Technology, Skolkovo Institute of Science and Technology (Skoltech) Moscow, Russia}
    \hspace{4pt}
    ~Andrei~Chertkov \footnotemark[1]
    \hspace{1pt}
    \footnotemark[2]
    \hspace{4pt}
    ~Roman~Schutski \footnotemark[2]
     \And
    ~Anh-Huy~Phan  \footnotemark[2]
    \hspace{4pt}
    ~Andrzej~Cichocki
    \footnotemark[2]  
    \hspace{1pt}
    \thanks{RIKEN Center for Advanced Intelligence Project (AIP), Tokyo, Japan} \thanks{Systems Research Institute, Polish Academy of Sciences, Warsaw, Poland}
    \hspace{4pt}
    ~Ivan~Oseledets 
    \footnotemark[2] 
    \hspace{1pt}
    \thanks{Artificial Intelligence Research Institute (AIRI), Moscow, Russia}
  }

\begin{document}

\maketitle

\begin{abstract}

We present a novel procedure for optimization based on the combination of efficient quantized tensor train representation and a generalized maximum matrix volume principle.
We demonstrate the applicability of the new Tensor Train Optimizer (TTOpt) method for various tasks, ranging from minimization of multidimensional functions to reinforcement learning.
Our algorithm compares favorably to popular gradient-free methods and outperforms them by the number of function evaluations or execution time, often by a significant margin.
\end{abstract}

\section{Introduction}
In recent years learning-based algorithms achieved impressive results in various applications, ranging from image and text analysis and generation~\cite{Ramesh2021ZeroShotTG} to sequential decision making and control~\cite{Mnih2015} and even quantum physics simulations~\cite{PhysRevResearch.2.033429}. 
The vital part of every learning-based algorithm is an optimization procedure, e.g., Stochastic Gradient Descent.
In many situations, however, the problem-specific target function is not differentiable, too complex, or its gradients are not helpful due to the non-convex nature of the problem~\cite{kolda2003ds, ALARIE2021100011}.
The examples include hyper-parameter selection during the training of neural models, policy optimization in reinforcement learning (RL), training neural networks with discrete (quantized) weights~\cite{Schweidtmann_2018} or with non-differentiable loss functions 
~\cite{NEURIPS2018_worldmodels}.
In all these contexts, efficient direct gradient-free optimization procedures are highly needed.

Recently,~\cite{Salimans2017EvolutionSA} showed that an essential class of gradient-free methods, namely the evolutionary strategies (ES)~\cite{Hansen2006,holland92, Schwefel1977}, are competitive in reinforcement learning problems. 
In RL, the goal is to find the agent's action distribution $\pi$ (the policy), maximizing some cumulative reward function $\fj$. 
The policy is usually parameterized with a set of parameters $\vt$.
It follows that the reward is a function of the parameters of the policy: $\fj(\pi(\vt)) = \fj(\vt)$.
The idea of~\cite{Salimans2017EvolutionSA} and similar works~\cite{NEURIPS2018_ARS,conti2018,self-guided-es,pmlr-v80-choromanski18a,NEURIPS2019_asebo} is to directly optimize the cumulative reward function $\fj(\vt) = \fj(\theta_1, \theta_2, \ldots, \theta_d)$ with respect to the parameters of the policy.
We pursued a similar approach to transform a traditional Markov Decision Process into an optimization problem (we provide the details in Appendix~B.3).
Although the agents trained with ES often demonstrate more rich behavior and better generalization compared to traditional gradient-based policy optimization, the convergence of ES is often slow~\cite{NEURIPS2019_asebo}.

As an alternative to previous works, we present a tensor-based\footnote{
    By tensors we mean multidimensional arrays with a number of dimensions $d$ ($d \geq 1$).
    A two-dimensional tensor ($d = 2$) is a matrix, and when $d = 1$ it is a vector.
    For scalars we use normal font, we denote vectors with bold letters and we use upper case calligraphic letters ($\tens{A}, \tens{B}, \tens{C}, \ldots$) for tensors with $d > 2$. Curly braces define sets.
    We highlight discrete and continuous scalar functions of multidimensional argument in the appropriate font, e.x., $\fj(\cdot)$,
    in this case, the maximum and minimum values of the function are denoted as $J_{max}$ and $J_{min}$, respectively.
} gradient-free optimization approach and apply it to advanced continuous control RL benchmarks. The algorithm that we called Tensor-Train (TT) Optimizer (\textbf{TTOpt}), works for multivariable functions with discrete parameters by reformulating the optimization problem in terms of tensor networks.
Consider a function $\fj(\vt): N^{d} \longrightarrow R$ of a $d$-dimensional argument $\vt$, where each entry $\theta_k$ of the vector $\vt$ takes a value in the discrete set $\{\omega_i\}_{i=1}^{N}$.
The function $\fj$ may be viewed as an implicitly defined $d$-dimensional tensor $\tj$. Each entry in $\tj$ is a value of $\fj$ for some argument.
Maximizing $\fj$ is equivalent to finding the sets of indices $\{\vt^{(m)}_{max}\}_{m=1}^{M}$ of maximal entries $\{j^{(m)}_{max}\}_{m=1}^{M}$ of $\tj$.

By using only a tiny fraction of adaptively selected tensor elements (e.g., a small number of function evaluations), 
our method builds a representation of $\tj$ in the TT-format and finds a set of the largest elements.
Although the algorithm works only with functions of discrete arguments, the grid size for each parameter can be huge thanks to the efficiency of the TT-format.
Fine discretization makes it possible to almost reach a continuous limit and obtain large precision with TTOpt.
The strength of our approach is, however, the direct handling of discrete (quantized) parameters.
\paragraph{Contributions.}
We propose an efficient gradient-free optimization algorithm for multivariable functions based on the low-rank TT-format and the generalized maximum matrix volume principle\footnote{
    We implemented the proposed algorithm within the framework of the publicly available software product: \url{https://github.com/AndreiChertkov/ttopt}.
}.
We demonstrate that our approach is competitive with a set of popular gradient-free methods for optimizing benchmark functions and neural network-based agents in RL problems. 
We empirically show that agents with discrete (heavily quantized) weights perform well in continuous control tasks. Our algorithm can directly train quantized neural networks, producing policies suitable for low-power devices.

\section{Optimization with tensor train}
\label{sec:ttopt}

{
In this section, we introduce a novel optimization algorithm.
We show how to represent the optimization problems in the discrete domain efficiently and then formulate optimization as a sampling of the objective function guided by the maximum volume principle. 
}

\subsection{Discrete formulation of optimization problems}

We first need to transfer the problem to the discrete domain to apply our method. It may seem that discretizing an optimization problem will make it harder. However, it will allow us to use powerful techniques for tensor network representation to motivate the algorithm.
For each continuous parameter $\theta_k$ ($k = 1, 2, \ldots, d$) of the objective function $\fj(\vt)$ we introduce a grid $\{\theta_{k}^{(n_k)}\}_{n_k=1}^{N_{k}}$.
At each point $(\theta_{1}^{(n_{1})}, \theta_{2}^{(n_{2})}, \ldots, \theta_{d}^{(n_{d})})$ of this grid with index $(n_{1}, n_{2}, \ldots, n_{d})$ the objective function takes a value
$
\fj(\theta_{1}^{(n_{1})}, \theta_{2}^{(n_{2})}, \ldots, \theta_{d}^{(n_{d})}) \equiv \tj[n_1, n_2, \ldots, n_d].
$
We thus can regard the objective function as an implicit $d$-dimensional tensor $\tj$ with sizes of the modes $N_{1}, N_{2}, \ldots, N_{d}$.
Finding the maximum of the function $\fj(\vt)$ translates into finding the maximal element of the tensor $\tj$ in the discrete setting.

Notice that the number of elements of $\tj$ equals:
$
\lvert \tj \rvert = N_1 \cdot N_2 \cdot \ldots \cdot N_d
\sim
\left(\max_{1 \leq k \leq d}{N_k}\right)^d.
$
The size of $\tj$ is exponential in the number of dimensions $d$. This tensor can not be evaluated or stored for sufficiently large $d$.
Fortunately, efficient approximations were developed to work with multidimensional arrays in recent years.
Notable formats include Tensor Train (TT)~\cite{oseledets2009breaking,oseledets2011tensor,phan2020a}, Tensor Chain/Tensor Ring~\cite{Khoromskij2011,8682231} and Hierarchical Tucker~\cite{hackbusch2009new}.
We use the most studied TT-format~\cite{CichockiBookPart1MAL059}, but the extensions of our method to other tensor decompositions are possible.

\subsection{Tensor Train decomposition}
{
\begin{definition}
A tensor $\tj \in \set{R}^{N_1 \times N_2 \times \cdots \times N_d}$ is said to be in the TT-format~\cite{oseledets2011tensor} if its elements are represented by the following expression
\begin{equation}\label{eq:tt_repr}
\tj [n_1, n_2, \ldots, n_d]
=
\sum_{r_0=1}^{R_0}
\sum_{r_1=1}^{R_1}
\cdots
\sum_{r_{d}=1}^{R_{d}}
    \tens{G}_1 [r_0, n_1, r_1]
    \tens{G}_2 [r_1, n_2, r_2]
    \ldots
    \tens{G}_d [r_{d-1}, n_d, r_d],
\end{equation}
where $n_k = 1, 2, \ldots, N_k$ for $k = 1, 2, \ldots, d$.
\end{definition}

In TT-format the $d$-dimensional tensor $\tj$ is approximated as a product of three-dimensional tensors $\tens{G}_k \in \set{R}^{R_{k-1} \times N_k \times R_k}$, called TT-cores.
The sizes of the internal indices $R_{0}, R_{1}, \cdots, R_{d}$ (with convention $R_{0} = R_{d} = 1$) are known as TT-ranks.
These ranks control the accuracy of the approximation.

The storage of the TT-cores, $\tens{G}_1, \tens{G}_2, \ldots, \tens{G}_d$, requires at most
$
d \cdot \max_{1 \leq k \leq d}{N_k}
\cdot \left(\max_{0 \leq k \leq d}{R_k}\right)^2
$
memory cells, and hence the TT-approximation is free from the curse of dimensionality\footnote{
    The number of elements of an uncompressed tensor (hence, the memory required to store it) and the number of elementary operations required to perform computations with such a tensor grow exponentially in dimensionality. This problem is called the curse of dimensionality.
} if the TT-ranks are bounded.
The basic linear algebra operations (such as finding a norm, differentiation, integration, and others) can also be implemented in the TT-format with polynomial complexity in dimensionality and mode size.
}
\paragraph{Building TT-approximation.}
Several efficient schemes were proposed to find TT-approximation if all or some of the elements of the initial tensor are known or may be generated by the function's call.
Examples include TT-SVD~\cite{oseledets2009breaking, oseledets2011tensor}, TT-ALS~\cite{holtz2012alternating} and TT-CAM~\cite{oseledets2010ttcross} (Cross Approximation Method in the TT-format).

We build upon the TT-CAM but modify it not to compute the approximation for the \emph{entire} tensor, but rather to find a \emph{small subset} of its maximal entries.
The original algorithm builds a TT-approximation by adaptively requesting elements of the input tensor.
As we will show below, these elements with high probability will have large absolute values.
Based on this observation, we formulate a robust optimization algorithm for multivariate functions (either discrete or continuous).
To simplify the understanding, we outline the approach for the two-dimensional case, and after that, we describe our gradient-free optimization method for the multidimensional case.

\subsection{Maximal element in a matrix}
{
The Cross Approximation Method (CAM) for matrices~\cite{goreinov2010find,CAIAFA2010557, ahmadi2021cross}
is a well-established algorithm for building a rank-$R$ approximation $\tilde{\matr{J}}$  of an implicitly given matrix $\matr{J}$:
\begin{equation}\label{eq:matr-cross}
\matr{J} \simeq \tilde{\matr{J}},
\quad
\tilde{\matr{J}}
=
\matr{J_C} \hat{\matr{J}}^{-1} \matr{J_R},
\end{equation}
where $\matr{J_C}$ consists of $R$ columns of $\matr{J}$, $\matr{J_R}$ is composed of $R$ rows of $\matr{J}$, and $\hat{\matr{J}}$ is a submatrix at their intersection.
Such approximation (also called cross or skeleton decomposition) may be built iteratively using a well known alternating directions method and a maximum volume (\func{maxvol}) algorithm\footnote{
    The \func{maxvol} algorithm finds $R$ rows in an arbitrary non-degenerate matrix $\matr{A} \in \set{R}^{N \times R}$ ($N > R$) which span a maximal-volume $R \times R$ submatrix $\hat{\matr{A}}$. The matrix $\hat{\matr{A}} \in \matr{A}$ has maximal value of the modulus of the determinant on the set of all nondegenerate square submatrices of the size $R \times R$.
    We describe the implementation of \func{maxvol} in Appendix~A.1.
    The algorithm greedily rearranges rows of $\matr{A}$ to maximize submatrix volume.
    Its computational complexity is $O(N R^2 + K N R)$, where $K$ is a number of iterations.
}~\cite{goreinov2010find}, as we will sketch below.
}

\paragraph{Intuition behind TTOpt.}
The main interest in optimization problems is not the approximation~\eqref{eq:matr-cross} itself, but the following property of the resulting maximum volume submatrix $\matr{\hat{J}}$.
\cite{goreinov2010find} proved that if $\matr{\hat{J}}$ is an $R \times R$ submatrix  of maximal volume (in selected rows and columns) then the maximal (by modulus) element $\hat{J}_{max} \in \matr{\hat{J}}$ bounds the absolute maximal element $J_{max}$ in the full matrix $\matr{J}$:
\begin{equation}
    \hat{J}_{max} \cdot R^2 \geq J_{max}.
    \label{eq:maxvol_maxelement}
\end{equation}
This statement is evident for $R=1$, and for the case $R > 1$ it gives an upper bound for the element.
By using elementwise transformations of $\matr{J}$, this upper bound can be used to obtain a sequence that converges to the global optimum. The main idea of the \func{maxvol}-based methods is that it is easier to find a submatrix with a large volume rather than the element with the largest absolute value. Moreover, our numerical experiments show that this bound is pessimistic, and in practice, the maximal-volume submatrix contains the element which is very close to the optimal one.


\begin{figure}
    \begin{minipage}{.5\textwidth}
        \vskip 0.6in
        \centering
        \includegraphics[scale=0.21]{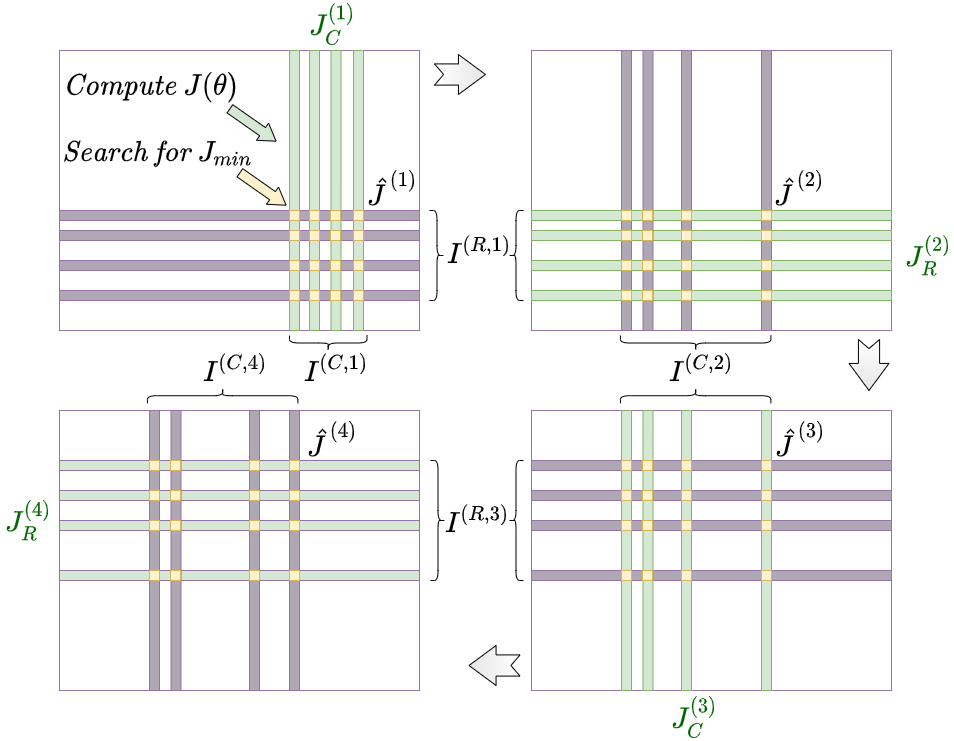}
        \vskip 0.11in
        \captionof{figure}{
            The scheme of the cross approximation algorithm for matrices using the alternating direction and maximal-volume principle. Green bars represent generated rows/columns; purple bars are rows/columns selected for generation in the next step by the \func{maxvol} algorithm.
            The method allows to find the optimum of the two dimensional function $\fj(\theta_{1}, \theta_{2}$).
        }
        \label{fig:cross_matrix}
    \end{minipage}
    \hspace{.045\textwidth}
    \begin{minipage}{.44\textwidth}
        \centering
        \includegraphics[scale=0.22]{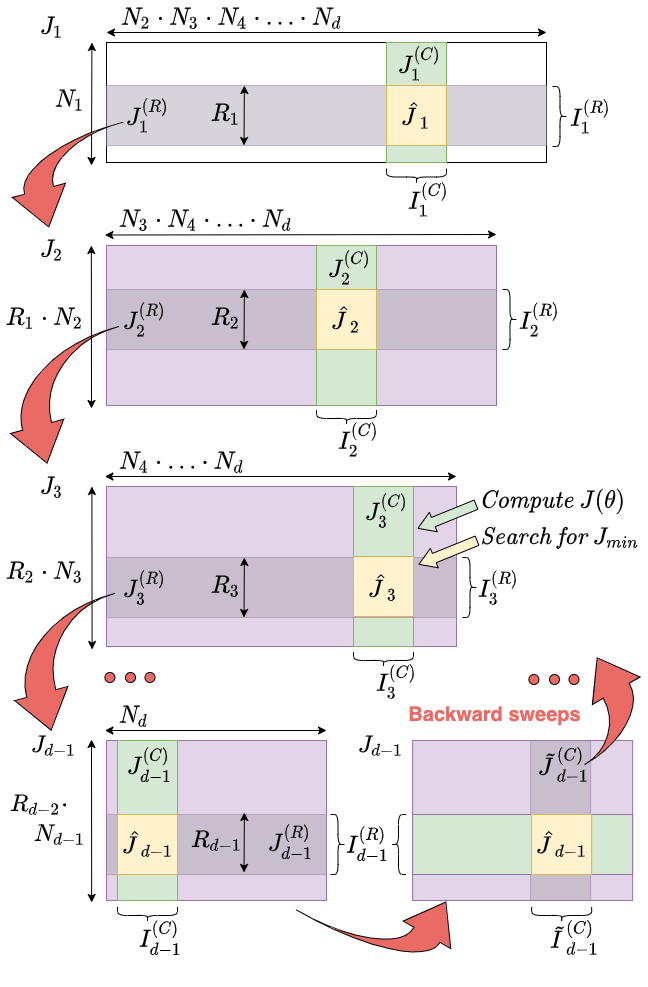}
        \captionof{figure}{
            Conceptual scheme of \func{TTOpt} algorithm based on the alternating direction and maximal-volume approaches for tensors.
            Only a small part of the tensor is explicitly generated during this procedure, as shown here with green columns.
            For the simplicity of presentation, the rows and columns selected at iterations are drawn as continuous blocks (they are not in practice).}
        \label{fig:cross_tt}
    \end{minipage}
\end{figure}

\paragraph{TTOpt algorithm for matrices.}
The idea of the TTOpt algorithm for matrices is to iteratively search for the maximal volume submatrices in the column and row space of the implicitly given\footnote{
    The matrix is specified as a function $\fj(\cdot)$ that allows to calculate the value of an arbitrary requested element $(n_1, n_2)$, where $1 \leq n_1 \leq N_1$ and $1 \leq n_2 \leq N_2$.
    We present the approach to approximate the value of the maximum modulus element of such a matrix.
    The method of finding the minimal or maximal elements within the framework of this algorithm will be described in Section~\ref{par:mapping_function}.
    
} input matrix $\matr{J} \in \set{R}^{N_1 \times N_2}$. After $T$ iterations a series of ``intersection'' matrices $\{ \matr{\hat{J}}^{(t)} \}_{t=1}^{T} \in \set{R}^{R \times R}$ is produced. The maximal element is searched in these small submatrices. We schematically represent the TTOpt algorithm in Figure~\ref{fig:cross_matrix}, and a description is given below:
\begin{enumerate}
    \item
        At the initial stage, we set the expected rank of the approximation, $R$, and select $R$ random columns $I^{(C,1)}$.
        We then generate the corresponding column submatrix $\matr{J_C}^{(1)} = \matr{J}[:, I^{(C,1)}] \in \set{R}^{N_1 \times R}$.
        Using the \func{maxvol} algorithm, we find the maximal-volume submatrix $\hat{\matr{J}}^{(1)} \in \set{R}^{R \times R}$ in $\matr{J_C}^{(1)}$ and store  its row indices in the list $I^{(R,1)}$.
    \item
        The indices $I^{(R,1)}$ are used to generate a row submatrix $\matr{J_R}^{(2)} = \matr{J}[I^{(R,1)}, :] \in \set{R}^{R \times N_2}$.
        Then, using the \func{maxvol} algorithm, we find the maximal-volume submatrix $\hat{\matr{J}}^{(2)}$ in the matrix $\matr{J_R}^{(2)}$ and store the corresponding column indices in the list $I^{(C, 2)}$.
    \item
        We generate the related columns $\matr{J_C}^{(3)} = \matr{J}[:, I^{(C, 2)}]$, apply again the \func{maxvol} algorithm to the column submatrix, and iterate the process until convergence.
    \item
        The approximate value of the maximum modulus element of the matrix $\matr{J}$ is found as
        \begin{equation}\label{eq:matrix-opt-value}
            \hat{J}_{max} = \max{\left(
                \,
                \max{(\hat{\matr{J}}^{(1)})}, \,
                \max{(\hat{\matr{J}}^{(2)})}, \,
                \ldots,
                \,
                \max{(\hat{\matr{J}}^{(T)})} \,
            \right)}.
        \end{equation}
\end{enumerate}

\subsection{Optimization in the multidimensional case}

As we explained previously, the target tensor, $\tj$, is defined implicitly, e.g., by a multivariable function $\fj$.
We propose a novel method to find the optimum in this implicit tensor.
We outline the approach below and provide detailed algorithms in Appendix~A.2.

As shown in Figure~\ref{fig:cross_tt}, we begin by considering the first unfolding\footnote{
    The $k$-th unfolding $\matr{J}_k$ for the $d$-dimensional tensor $\tj \in \set{R}^{N_1 \times N_2 \times \cdots \times N_d}$ is the matrix $
        \matr{J}_k \in
            \set{R}^{N_1 \ldots N_k \times N_{k+1} \ldots N_d},
        $ with elements
        $
        \matr{J}_k[\,
            \overline{n_1, \ldots, n_k},
            \overline{n_{k+1}, \ldots, n_d}
        \,] \equiv \tj[n_1, n_2, \ldots, n_d]
        $
        for all indices.
} $\matr{J}_1 \in \set{R}^{N_1 \times N_2 \ldots N_d}$ of the tensor $\tj$ and select $R_1$ random columns $I_1^{(C)}$. Precisely, $I_1^{(C)}$ here is a list of $R_1$ random multi-indices of size $d-1$, which specify positions along modes $k=2$ to $k=d$.
We then generate the submatrix $\matr{J}_1^{(C)} \in \set{R}^{N_1 \times R_1}$ for all positions along the first mode (shown in green in Figure~\ref{fig:cross_tt}). Like in matrix case, we apply the \func{maxvol} algorithm to find the maximal-volume submatrix $\hat{\matr{J}_1} \in \set{R}^{R_1 \times R_1}$ and store the corresponding indices of $R_1$ rows in the list $I_1^{(R)}$.

In contrast with matrix case, we cannot generate the row submatrix $\matr{J}_{1}^{(R)} \in \set{R}^{R_1 \times N_2 N_3 \ldots N_d}$ for the selected row indices $I_1^{(R)}$, since it contains an exponential number of elements. 
The following trick is used instead. We consider the implicit matrix $\matr{J}_{1}^{(R)}$ and reshape it to a new matrix $\matr{J}_{2} \in \set{R}^{R_1 N_2 \times N_3 \ldots N_d}$.
We sample $R_2$ random columns $I_2^{(C)}$ in the matrix $\matr{J}_{2}$ and generate the entire small submatrix  $\matr{J}_2^{(C)} \in \set{R}^{R_1 N_2 \times R_2}$.
Next we find the maximal-volume submatrix $\hat{\matr{J}}_2 \in \set{R}^{R_2 \times R_2}$ in $\matr{J}_2^{(C)}$ and store the corresponding $R_2$ row multi-indices in the list $I_2^{(R)}$.
The resulting submatrix $\matr{J}_2^{(R)} \in \set{R}^{R_2 \times N_3 N_4 \ldots N_d}$ is then transformed (without explicitly evaluating its elements) into
the matrix $\matr{J}_{3} \in \set{R}^{R_2 N_3 \times N_4 \ldots N_d}$.


We continue the described operations, called sweeps, until the last mode of the initial tensor, $\tens{J}$, is reached.
After that, we repeat the process in the opposite direction, sampling now the row indices instead of the column indices.
These sequences of forward and backward sweeps are continued until the algorithm converges to some row and column indices for all unfolding matrices\footnote{
    The TT-approximation~\eqref{eq:tt_repr} of the tensor $\tens{J}$ may be recovered from the generated columns $\matr{J}_k^{(C)}$ and maximal-volume submatrices $\hat{\matr{J}}_k$ ($k = 1, 2, \ldots, d$) as follows: $\matr{G}_k = \matr{J}_k^{(C)} \hat{\matr{J}}_k^{-1} \in \set{R}^{R_{k-1} N_k \times R_k}$, where $\matr{G}_k$ is the 2-th unfolding of the $k$-th TT-core.
    However, we do not consider this point in more detail, since in our work, the main task is to find the minimum or maximum value of the tensor, not to construct its low-rank approximation.
} or until the user-specified limit on the number of requests to the objective function, $\fj$, is exceeded.
Finally, after $T$ sweeps, the approximate value of the maximum modulo element can be found by the formula~\eqref{eq:matrix-opt-value}, as in the two-dimensional case.

Note that currently there is no analog of Eq.~\eqref{eq:maxvol_maxelement} in the multidimensional case, and hence there are no formal guarantees of the convergence of the sweeps to the global minimum, nor the rate of this convergence. The only guarantee is that the result will monotonically improve with iterations.

\subsection{Complexity of the algorithm}
It can be easily shown that the described algorithm requires to evaluate only $\order{d \cdot \max_{1 \leq k \leq d}{\left(N_k R_k^2\right)}}$ elements of the implicit tensor in one sweep.
Thus, with a total number of sweeps $T$, we will have
\begin{equation}\label{eq:ttopt-evaluations}
\order{T \cdot d \cdot \max_{1 \leq k \leq d}{\left(N_k R_k^2\right)}},
\end{equation}
calls to the objective function $\fj$.
In practice, it turns out to be more convenient to limit the maximum number of function calls, $M$, according to the computational budget.

If the time of a single call to $\fj$ is significant, then the effort spent on the algorithm's operation will be negligible.
Otherwise, the bottleneck will be the calculation of the maximal-volume submatrices by the \func{maxvol} algorithm.
Taking into account the estimate of the \func{maxvol} complexity, given in Appendix~A.1, it can be shown that in this case the complexity of our algorithm is
\begin{equation}\label{eq:ttopt-complexity}
\order{T \cdot d \cdot \max_{1 \leq k \leq d}{\left(N_k R_k^3\right)}}.
\end{equation}

\subsection{Implementation details}
\label{s:details}
{
For the effective implementation of the TTOpt algorithm, the following important points should be taken into account (see also the detailed pseudocode in Appendix~A.2).
}

\paragraph{Stability.}
Submatrices $\matr{J}_{k}^{(C)}$ (or $\matr{J_C}^{(k)}$ and $\matr{J_R}^{(k)}$ for the two-dimensional case) that arise during the iterations may degenerate, and in this case it is impossible to apply the \func{maxvol} algorithm.
To solve this problem, we first calculate the QR decomposition for these matrices and then apply the \func{maxvol} to the corresponding $\matr{Q}$ factors\footnote{
    It can be shown that this operation does not increase the complexity estimate~\eqref{eq:ttopt-complexity} of the algorithm.
}.
\paragraph{Rank selection.}
We do not know in advance the exact ranks $R_1, R_2, \ldots, R_d$ of the unfolding matrices (or rank $R$ of the matrix $\matr{J}$ for the two-dimensional case).
Therefore, instead of the \func{maxvol} algorithm, we use its modification, i.e., the \func{rect\_maxvol} algorithm\footnote{
    The \func{rect\_maxvol} algorithm allows to find $R + \Delta R$ rows in an arbitrary nondegenerate matrix $\matr{A} \in \set{R}^{N \times R}$ ($N > R$, $N \geq R + \Delta R$) which form an approximation to the \emph{rectangular} maximal-volume submatrix of the given matrix $\matr{A}$.
    Details about \func{rect\_maxvol} are provided in Appendix~A.1.
}~\cite{mikhalev2018rectangular}, within the framework of which several (``most important'') rows are added to the maximal-volume submatrix.
In this case, we have $\hat{\matr{J}}_{1}^{(C)} \in \set{R}^{(R_1 + \Delta R_1) \times R_1}$, $\hat{\matr{J}}_{2}^{(C)} \in \set{R}^{(R_2 + \Delta R_2) \times R_2}$, etc.
Note that the final approximation obtained in this case may have an overestimated rank, which, if necessary, can be reduced by appropriate rounding, for example, by truncated SVD decomposition.

\paragraph{Mapping function.\label{par:mapping_function}}
Maximal-volume submatrices contain the maximum modulus element but not the minimum or maximum element of the tensor (i.e., the sign is not taken into account).
We introduce a dynamic mapping function to find the global minimum (or maximum). Instead of $\fj$ at each step of the algorithm, we evaluate\footnote{
    The mapping function should be continuous, smooth, and strictly monotone.
    There are various ways to choose such a function. However, during test runs, it turned out that the proposed function~\eqref{eq:ttopt-map} is most suitable.
}:
\begin{equation}\label{eq:ttopt-map}
\func{g}(\vx)
=
\frac{\pi}{2} -
\func{atan}\left(
    \func{J}(\vect{x}) - J_{min}
\right),
\end{equation}
where $J_{min}$ is the current best approximation for the minimum element of the tensor.

\paragraph{Quantization.}
To reach high accuracy, we often need fine grids. In the case when the sizes of the tensor modes $N_1, N_2, \ldots, N_d$ are large, the sizes of unfolding matrices become large, which leads to a significant increase in computational complexity of the \func{maxvol} algorithm.
To solve this problem, we apply additional compression based on quantization of the tensor modes~\cite{oseledets2010approximation}.
Assume without the loss of generality that the size of each mode is $N_k = P^{q}$ ($k = 1, 2, \ldots, d$; $P \geq 2$; $q \geq 2$).
Then we can reshape the original $d$-dimensional tensor $\tj \in \set{R}^{N_1 \times N_2 \times \cdots \times N_d}$ into the tensor $\tilde{\tj} \in \set{R}^{P \times P \times \cdots \times P}$ of a higher dimension $d \cdot q$, but with smaller modes of size $P$.
The \func{TTOpt} algorithm can be applied for this ``long'' tensor instead of the original one\footnote{
    If the maximal rank $R > P$, we select all indices for first $k$ modes until $P^k > R$. We note however that this detail does not change the global behavior of the TTOpt algorithm.
}.
We found that this idea significantly boosts the accuracy of our algorithm and reduces the complexity and execution time.
Typically, $P$ is taken as small as possible, e.g. $P = 2$.


\section{Experiments}
\label{sec:exp}

To demonstrate the advantage of the proposed optimization method, we tested \func{TTOpt} on several numerical problems.
First, we consider analytical benchmark functions and then the practically significant problem of optimizing the parameters of the RL agent.
We select
\func{GA} (Genetic Algorithm~\cite{holland92,SimpleGASuch2017-DeepNG}),
\func{openES} (basic version of OpenAI Evolution Strategies~\cite{Salimans2017EvolutionSA}) and
\func{cmaES} (the Covariance Matrix Adaptation Evolution Strategy~\cite{Hansen2006})
as baselines for both experiments.
Additionally we used
\func{DE} (Differential Evolution~\cite{Storn1997-DE}),
\func{NB} (NoisyBandit method from Nevergrad~\cite{Nevergrad}) and
\func{PSO} (Particle Swarm Optimization~\cite{HEIN201787,10.1016/j.engappai.2020.103525})
for benchmark functions\footnote{
    We used implementations of the methods from available packages \func{estool}~(\url{https://github.com/hardmaru/estool}),
    \func{pycma}~(\url{https://github.com/CMA-ES/pycma}, and
    \func{nevergrad}~(\url{https://github.com/facebookresearch/nevergrad}).
}.
We also compared the proposed approach with gradient-based methods applied for all benchmark functions\footnote{
    We used implementations from the package \url{https://github.com/rfeinman/pytorch-minimize}).
    We carried out computations with all methods from this library, except for Trust-Region GLTR (Krylov) and Dogleg methods, for which the calculation ended with an error for most benchmarks.
}:
\func{BFGS} (Broyden–Fletcher–Goldfarb–Shanno algorithm),
\func{L-BFGS} (Limited-memory BFGS),
\func{CG} (Conjugate Gradient algorithm),
\func{NCG} (Newton CG algorithm),
\func{Newton} (Newton Exact algorithm),
\func{TR NCG} (Trust-Region NCG algorithm) and
\func{TR} (Trust-Region Exact algorithm).

According to our approach, the \func{TTOpt} solver has the following configurable parameters: $\bf{a}$ and $\bf{b}$ are lower and upper grid bounds (for simplicity, we use the same value for all dimensions); $\bf{R}$ is a rank (for simplicity, we use the same value for all unfolding matrices); $\bf{P}$ is a submode size (mode size of the quantized tensor; for simplicity, we use the same value for all dimensions); $\bf{q}$ is the number of submodes in the quantized tensor (each mode of the original tensor has size $N = P^q$); $\bf{M}$ is a limit on the number of requests to the objective function.

\begin{table*}
\caption{
    Comparison of the \func{TTOpt} optimizer versus baselines in terms of the final error $\epsilon$ (absolute deviation of the obtained optimal value relative to the global minimum) and computation time $\tau$ (in seconds) for various benchmark functions.
    See Table~1 in Appendix~B.1 with the list of functions and their properties.
    The reported values are averaged over 10 independent runs. 
    A upper half of the table presents gradient free  (zeroth order) methods, and lower half is for first and second order methods.
}

\begin{center}
\begin{tiny}
\begin{sc}

\begin{tabular}{|p{0.99cm}|p{0.1cm}|p{0.78cm}|p{0.78cm}|p{0.78cm}|p{0.78cm}|p{0.78cm}|p{0.78cm}|p{0.78cm}|p{0.78cm}|p{0.78cm}|p{0.78cm}|}\hline

&
& \emph{F1}
& \emph{F2}
& \emph{F3}
& \emph{F4}
& \emph{F5}
& \emph{F6}
& \emph{F7}
& \emph{F8}
& \emph{F9}
& \emph{F10} \\ \hline


\multirow{2}{*}{\func{TTOpt}}
& $\epsilon$ & \textbf{ 3.9e-06} & \textbf{ 2.9e-07} &  1.8e-12 & 4.4e-15 &  2.8e-02 & \textbf{ 1.1e-01} & 5.5e-09 & \textbf{ 4.6e-11} &  1.8e-01 & \textbf{ 1.3e-04} 
\\
& $\tau$ & \textit{  2.61} & \textit{  2.44} & \textit{  2.45} & \textit{  2.52} & \textit{  2.40} & \textit{  2.48} & \textit{  2.39} & \textit{  2.60} & \textit{  2.32} & \textit{  2.44}  \\ \hline 

\multirow{2}{*}{\func{GA}}
& $\epsilon$ &  9.7e-02 &  8.4e-03 &  5.8e-03 &  2.0e+00 & \textbf{ 3.9e-04} &  1.0e+01 &  1.2e-01 &  7.9e-01 & \textbf{ 6.2e-03} &  4.2e+03 
\\
& $\tau$ & \textit{  6.21} & \textit{  4.56} & \textit{  5.09} & \textit{  4.87} & \textit{  5.85} & \textit{  5.69} & \textit{  5.05} & \textit{  5.04} & \textit{  5.04} & \textit{  4.70}  \\ \hline 

\multirow{2}{*}{\func{openES}}
& $\epsilon$ &  1.8e-01 &  1.2e-02 &  1.7e-02 &  2.0e+00 &  1.2e-03 &  9.7e+00 &  3.8e+00 &  2.1e+00 &  1.8e-02 &  4.2e+03 
\\
& $\tau$ & \textit{  2.62} & \textit{  1.08} & \textit{  1.62} & \textit{  1.08} & \textit{  2.41} & \textit{  2.04} & \textit{  1.30} & \textit{  1.39} & \textit{  1.62} & \textit{  1.12}  \\ \hline 

\multirow{2}{*}{\func{cmaES}}
& $\epsilon$ & 5.1e+287 &  3.1e-01 & \textbf{ 9.3e-77} &  2.0e+00 & 7.6e+289 &  1.9e+01 &  5.5e-02 &  9.3e+01 & 5.3e+289 & 1.7e+282 
\\
& $\tau$ & \textit{ 10.36} & \textit{  8.50} & \textit{  9.40} & \textit{  9.13} & \textit{ 12.86} & \textit{  9.76} & \textit{  9.04} & \textit{  9.13} & \textit{ 11.15} & \textit{  8.86}  \\ \hline 

\multirow{2}{*}{\func{DE}}
& $\epsilon$ &  1.1e+00 &  4.3e-02 &  3.3e-02 &  9.0e-05 &  1.8e-01 &  2.6e-01 &  2.0e-01 &  6.2e+00 &  6.6e-01 &  3.8e+02 
\\
& $\tau$ & \textit{ 38.91} & \textit{ 38.06} & \textit{ 51.05} & \textit{ 39.48} & \textit{ 41.35} & \textit{ 41.40} & \textit{ 41.34} & \textit{ 41.31} & \textit{ 37.97} & \textit{ 38.64}  \\ \hline 

\multirow{2}{*}{\func{NB}}
& $\epsilon$ &  1.5e+01 &  6.5e+00 &  3.9e+01 &  1.2e-01 &  2.4e+01 &  6.6e+00 &  2.6e+10 &  6.3e+01 &  3.4e+00 &  3.2e+03 
\\
& $\tau$ & \textit{ 45.23} & \textit{ 46.98} & \textit{ 37.50} & \textit{ 45.91} & \textit{ 48.03} & \textit{ 37.16} & \textit{ 40.05} & \textit{ 44.95} & \textit{ 44.06} & \textit{ 46.91}  \\ \hline 

\multirow{2}{*}{\func{PSO}}
& $\epsilon$ &  1.2e+01 &  5.3e+00 &  3.5e+01 &  9.8e-02 &  2.0e+01 &  2.5e-01 &  2.0e+10 &  2.3e+01 &  5.1e-01 &  2.9e+03 
\\
& $\tau$ & \textit{ 47.19} & \textit{ 47.04} & \textit{ 45.50} & \textit{ 43.39} & \textit{ 46.80} & \textit{ 44.97} & \textit{ 46.46} & \textit{ 42.78} & \textit{ 43.15} & \textit{ 47.13}  \\ \hline  \hline 



\multirow{2}{*}{\func{BFGS}}
& $\epsilon$ &  1.9e+01 &  2.1e+00 &  1.9e+01 &  4.3e-13 &  1.2e-02 &  6.4e+00 &  2.4e-05 &  7.0e+01 &  4.2e+00 &  2.1e+03 
\\
& $\tau$ & \textit{  0.01} & \textit{  0.02} & \textit{  0.03} & \textit{  0.00} & \textit{  0.02} & \textit{  0.01} & \textit{  0.04} & \textit{  0.01} & \textit{  0.01} & \textit{  0.00}  \\ \hline 

\multirow{2}{*}{\func{L-BFGS}}
& $\epsilon$ &  1.9e+01 &  1.9e+00 &  4.0e-10 &  4.3e-13 &  1.2e-02 &  4.5e+00 &  4.5e-10 &  7.0e+01 &  4.2e+00 &  2.1e+03 
\\
& $\tau$ & \textit{  0.01} & \textit{  0.06} & \textit{  0.05} & \textit{  0.00} & \textit{  0.01} & \textit{  0.02} & \textit{  0.02} & \textit{  0.01} & \textit{  0.01} & \textit{  0.01}  \\ \hline 

\multirow{2}{*}{\func{CG}}
& $\epsilon$ &  1.9e+01 &  3.4e+00 & N/A &  \textbf{0.0e+00} &  2.0e-02 &  4.4e+00 &  2.9e-12 &  7.0e+01 &  4.2e+00 &  2.1e+03 
\\
& $\tau$ & \textit{  0.01} & \textit{  0.07} & N/A & \textit{  0.00} & \textit{  0.01} & \textit{  0.05} & \textit{  0.02} & \textit{  0.01} & \textit{  0.03} & \textit{  0.00}  \\ \hline 

\multirow{2}{*}{\func{NCG}}
& $\epsilon$ &  1.9e+01 &  3.4e+00 &  1.7e-19 &  \textbf{0.0e+00} &  7.4e-02 &  6.4e+00 &  2.1e-12 &  7.0e+01 &  4.2e+00 &  2.2e+03 
\\
& $\tau$ & \textit{  0.01} & \textit{  0.06} & \textit{  0.04} & \textit{  0.00} & \textit{  0.01} & \textit{  0.06} & \textit{  0.01} & \textit{  0.00} & \textit{  0.02} & \textit{  0.01}  \\ \hline 

\multirow{2}{*}{\func{Newton}}
& $\epsilon$ &  1.9e+01 &  2.2e+00 &  1.1e+11 &  \textbf{0.0e+00} &  3.2e-02 &  6.4e+00 &  \textbf{2.8e-24} &  7.0e+01 &  4.2e+00 &  2.1e+03 
\\
& $\tau$ & \textit{  0.01} & \textit{  0.02} & \textit{  0.01} & \textit{  0.00} & \textit{  0.01} & \textit{  0.10} & \textit{  0.01} & \textit{  0.01} & \textit{  0.01} & \textit{  0.00}  \\ \hline 

\multirow{2}{*}{\func{TR NCG}}
& $\epsilon$ &  1.9e+01 &  6.5e+00 &  2.4e-10 &  4.0e-12 &  4.8e+00 &  4.4e+00 &  1.2e-14 &  7.0e+01 &  4.2e+00 &  2.1e+03 
\\
& $\tau$ & \textit{  0.01} & \textit{  0.12} & \textit{  0.04} & \textit{  0.00} & \textit{  0.02} & \textit{  0.03} & \textit{  0.01} & \textit{  0.00} & \textit{  0.02} & \textit{  0.01}  \\ \hline 

\multirow{2}{*}{\func{TR}}
& $\epsilon$ &  1.9e+01 &  7.0e+00 &  3.2e-13 &  4.0e-12 &  6.1e+01 &  2.6e+00 &  5.3e-11 &  7.0e+01 &  4.2e+00 &  2.1e+03 
\\
& $\tau$ & \textit{  0.02} & \textit{ 12.97} & \textit{  0.06} & \textit{  0.00} & \textit{  0.06} & \textit{  0.05} & \textit{  0.02} & \textit{  0.01} & \textit{  0.01} & \textit{  0.01}  \\ \hline 


\end{tabular}
\end{sc}
\end{tiny}
\end{center}
\label{tab:compare_benchmark_result}
\end{table*}

\begin{table}
\caption{The result of the \func{TTOpt} optimizer in terms of the final error $\epsilon$ (absolute deviation of the obtained optimal value relative to the global minimum) and computation time $\tau$ (in seconds) for  benchmark \emph{F1} (Ackley function) for various dimension numbers.}
\vskip 0.05in

\begin{center}
\begin{small}
\begin{sc}

\begin{tabular}{|p{1.85cm}|p{2.5cm}|p{2.5cm}|p{2.5cm}|p{2.5cm}|}\hline

Dimension       &
$d \, = \,  10$ &
$d \, = \,  50$ &
$d \, = \, 100$ &
$d \, = \, 500$ \\ \hline

Error, $\epsilon$
& 3.9e-06
& 3.9e-06
& 3.9e-06
& 3.9e-06
\\

Time, $\tau$
& \textit{3.1}
& \textit{40.1}
& \textit{143.9}
& \textit{3385.3}
\\ \hline

\end{tabular}
\end{sc}
\end{small}
\end{center}
\label{tab:compare_benchmark_dim}
\end{table}

\subsection{Benchmark functions minimization}
\label{sec:exp_benchmark}

To analyze the effectiveness of the \func{TTOpt}, we applied it to $10$-dimensional benchmark functions with known global minimums; see Table~1 in Appendix~B.1 with the list of functions and their properties (note that some of the considered benchmarks are multimodal non-separable functions).
Also, in Appendix~B.1, we present a more detailed study of the \func{TTOpt} solver and the dependence of the accuracy on the value of its parameters ($R$, $q$ and $M$).

In all experiments with baselines (\func{GA}, \func{openES}, \func{cmaES}, \func{DE}, \func{NB}, \func{PSO}, \func{BFGS}, \func{L-BFGS}, \func{CG}, \func{NCG}, \func{Newton}, \func{TR NCG} and \func{TR}), we used default parameter values.
In Appendix~B.2 we also present the additional experiments with Bayesian Optimization~\cite{JMLR:v22:18-220}.
For \func{TTOpt} we selected rank\footnote{
    We chose rank $R$ using the following heuristic. The minimal number of sweeps is fixed as $T=5$. It follows that the algorithm will need  $ 2 \cdot T \cdot (d q) \cdot P \cdot R^2$ function calls. With a given limit on the number of function requests $M$, the rank can be estimated as $R \leq \sqrt{\frac{M}{2 \cdot T \cdot d \cdot q \cdot P}}$.
} $R=4$, submode size $P=2$ and the number of submodes $q=25$.
For all methods, a limit on the number of requests to the objective function is chosen as $M = 10^{5}$.
All calculations are performed on a standard laptop.

The results are demonstrated in Table~\ref{tab:compare_benchmark_result}.
For each method we list the absolute deviation of the result $\hat{J}_{min}$ from the global minimum $J_{min}$, i.e., $\epsilon = |\, \hat{J}_{min} - J_{min} \,|$. We also present the total running time, $\tau$.
Compared to other benchmarks, \func{TTOpt} is consistently fast, accurate and avoids random failures to converge seen in other algorithms.
Additionally, \func{TTOpt} turns out one of the fastest gradient-free algorithms (\func{GA}, \func{openES}, \func{cmaES}, \func{DE}, \func{NB}, \func{PSO}), despite a simple Python implementation.

One of the advantages of the proposed \func{TTOpt} approach is the possibility of its application to essentially multidimensional functions.
In Table~\ref{tab:compare_benchmark_dim} we present the result of \func{TTOpt} for the \emph{F1} benchmark function of various dimensionality (results for other benchmarks are in Appendix~B.1).
Note that as a limit on the number of requests to the objective function we chose $M = 10^4 \cdot d$, and the values of the remaining parameters were chosen the same as above.
As can be seen, even for $500$-dimensional functions, the \func{TTOpt} method gives a fairly accurate result.

\subsection{Application of TTOpt to Reinforcement Learning}

\begin{table}[t!]
\vskip -0.12in
\caption{Mean $\mathbb{E}$ and standard deviation $\sigma$ of the final cumulative reward. The environments are encoded using capital letters \textbf{S}wimmer-v3, \textbf{L}unarLanderContinuous-v2, \textbf{I}nvertedPendulum-v2 and \textbf{H}alfCheetah-v3. The left sub-table is for mode size $N=3$, another one is for mode size $N=2^{8}$. All runs are averaged over seven random seeds.
}
\vskip 0.15in
\begin{center}
\begin{small}
\begin{sc}
\label{tab:exp_31}
\label{tab:exp_qtt28}
\begin{tabular}{|p{1.2cm}|p{0.4cm}|p{1cm}|p{1cm}| p{1cm}|p{1cm}|
p{1cm}|p{1cm}|p{1cm}|p{1cm}|p{1cm}|}

\hline
{} & {} & $S(3^1)$ & $L(3^1)$ & $I(3^1)$ & $H(3^1)$  & $S(2^8)$ & $L(2^8)$ &  $I(2^8)$ & $H(2^8)$\\ \hline

\multirow{2}{*}{\func{TTOpt}} &
$\mathbb{E}$\newline$\sigma$  &   \textbf{357.50\newline6.59} &   \textbf{290.29\newline24.40} &   \textbf{1000.00\newline0.00} &  \textbf{4211.02} \newline 211.94 &  311.82\newline29.61 &   \textbf{286.87\newline21.65} &   \textbf{1000.00\newline0.00} &  2935.90\newline544.11 \\ \hline

\multirow{2}{*}{\func{GA}} &
$\mathbb{E}$\newline$\sigma$   &  349.91\newline10.04 &   283.05\newline16.28 &  893.00\newline283.10 &  2495.37\newline185.11 &   \textbf{359.79\newline4.21} &   213.75\newline99.67 &  222.86\newline342.79 &  \textbf{3085.80\newline842.76} \\ \hline

\multirow{2}{*}{\func{cmaES}} &
$\mathbb{E}$\newline$\sigma$ &      342.31\newline36.07 &   214.55\newline93.79 &  721.00\newline335.37 &  2549.83\newline501.08 &  340.54\newline78.90 &  221.95\newline133.80 &  621.00\newline472.81 &  2879.46\newline929.55 \\ \hline

\multirow{2}{*}{\func{openES}} &
$\mathbb{E}$\newline$\sigma$ &   318.39\newline44.61 &  114.97\newline113.48 &  651.86\newline436.37 &  2423.16\newline602.43 &  109.39\newline40.11 &   73.08\newline163.33 &  224.71\newline217.51 &  1691.22\newline976.96 \\ \hline

\end{tabular}
\end{sc}
\end{small}
\end{center}
\vskip -0.2in
\end{table}

We used several continuous RL tasks implemented in Mujoco~\cite{mujoco} and OpenAI-GYM~\cite{Brockman2016OpenAIG}: Swimmer-v3~\cite{Coulom-2002a}, LunarLanderContinuous-v2, InvertedPendulum-v2 and HalfCheetah-v3~\cite{4400335}.
In all experiments, the policy $\pi$ is represented by a neural network with three hidden layers and with \func{tanh} and \func{ReLU} activations. Each layer is a convolution layer. See additional details about hyperparameters in Table~6 in Appendix~B. 

We discretize (quantize) the values of agent's weights. The \func{TTOpt} method is used to optimize discrete agent's weights in order to maximize the cumulative reward of the episode.
This corresponds to on-policy learning.

To properly compare \func{TTOpt} with other methods, we propose modified evolutionary baselines that enforce constrained parameter domain.
We adapt penalty term and projection techniques from~\cite{constraint-es1,constraint-es2} to introduce constraints, see Appendix~B.4 for details.


First, we run benchmarks with small mode size $N=3^{1}$ with lower and upper grid bounds $\pm 1$. Another series of experiments was done with
finer mode of size $N=2^{q}$ with the same bounds. These experiments model the case of neural networks with discrete (quantized) weights which use $q$-bits quantization.
Finally, we provide the results of using \func{TTOpt} as a fine-tuning procedure for linear policies from~\cite{NEURIPS2018_ARS}.

We present characteristic training curves based on the number of environment interactions and execution time for the HalfCheetah-v3 experiment ($N=3$) in Figure~\ref{fig:exp_1}.
Training curves for other environments can be found in Appendix~B.5, in Figure~3 for $N=3$ and in Figure~4 for $N=256$.
\func{TTOpt} consistently outperforms all other baselines on the coarse grid with mode size $N=3$.
Our method is still best for finer grids with mode size $N=256$ on InvertedPendulum-v2 and LunarLanderContinuous-v2, and second-best on HalfCheetah-v3.
Moreover, \func{TTOpt} has significantly lower execution time compared to evolutionary baselines.
Another interesting observation is that the training curves of \func{TTOpt} have low dispersion, e.g., the algorithm performs more consistently than the baselines (see Appendix~B.5). 
Table~\ref{tab:exp_31} summarizes the experiments for the coarse and fine grids. Results for fine-tuning of linear policies are presented in Table~5 in Appendix~B.5. We also did rank and reward dependency study in Appendix~B.

\begin{figure}[t!]
    \centering
    \includegraphics[scale=0.09]{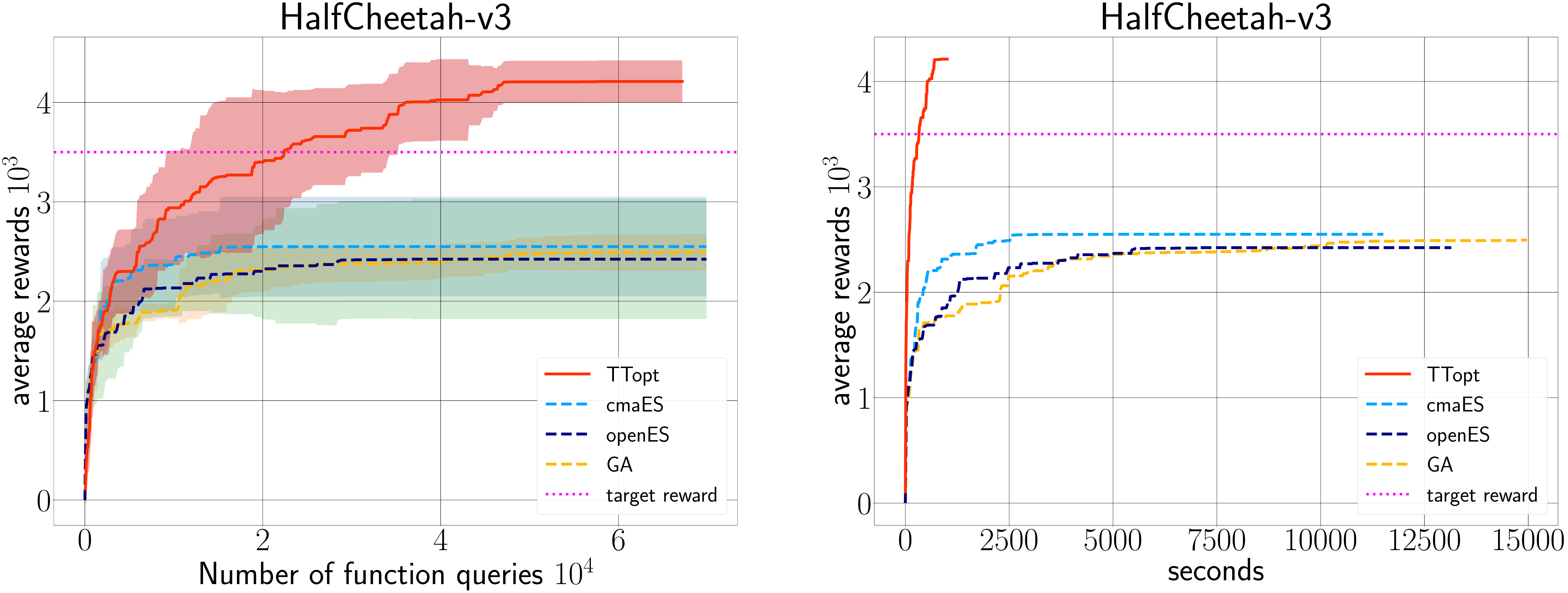}
    \caption{Training curves of \func{TTOpt} and baselines for HalfCheetah-v3 ($N=3$). The upper plot is the average cumulative reward versus the number of interactions with the environment. The lower plot is the same versus execution time. The reward is averaged for seven seeds. The shaded area shows the difference of one standard deviation around the mean. See similar plots for other environments in Appendix~B.5.
    }
    \label{fig:exp_1}
\end{figure}



\section{Related work}
\label{sec:related}

In the case of high dimensional optimization, evolutionary strategies (ES)~\cite{Schwefel1977, Nesterov2017} are one of the most advanced methods of black-box optimization.
This approach aims to optimize the parameters of the search distribution, typically a multidimensional Gaussian, to maximize the objective function. Finite difference schemes are commonly used to approximate gradients of the parameters of the search distribution. Numerous works proposed techniques to improve the convergence of ES~\cite{Nesterov2017}. 
\cite{JMLR:v15:wierstra14a} proposed to use second-order natural gradient of~\cite{blind} to generate updates, while ~\cite{Hansen2006} suggested to include the history of recent updates to generate next ones. ~\cite{pmlr-v97-maheswaranathan19a} presented the concept of surrogate gradients for faster convergence.

Another series of works aimed to reduce the high sampling complexity of ES. In ~\cite{NEURIPS2019_asebo} the authors described how to use active subspaces ~\cite{ActiveSubspace} to reduce the number of objective function calls dynamically.

Plenty of the already mentioned works in gradient-free optimization specifically applied these methods to RL tasks ~\cite{Salimans2017EvolutionSA, pmlr-v80-choromanski18a,NEURIPS2019_asebo,NEURIPS2018_ARS,conti2018,self-guided-es,HEIDRICHMEISNER2009152}.
Overall, the performance of ES-based methods is comparable to conventional policy gradients, especially if the number of model parameters is small~\cite{trpo-pmlr-v37-schulman15,ppo-Schulman2017ProximalPO}. Another advantage of ES over policy gradient is that it produces more robust and diverse policies~\cite{es-not-fd-approx,self-guided-es} by eliminating the problem of delayed rewards and short length time horizons. Finally, evolutionary approaches are suitable for the problems with non-Markovian properties~\cite{HEIDRICHMEISNER2009152}.

Other metaheuristic~\cite{HerrmannPriceJoyce+2020+45+62,9524335} and classical optimization~\cite{pmlr-v32-qin14} techniques are also studied within RL scope. The examples include simulated annealing~\cite{1562233}, particle swarm optimization~\cite{HEIN201787,10.1016/j.engappai.2020.103525} and even classical Nelder-Mead algorithm~\cite{NeldMead65, 7379496}. These methods, however, are not tested on common RL task sets. Several other works combined evolutionary methods with RL to achieve better performance in complex scenarios~\cite{10.5555/3326943.3327053,Faust2019EvolvingRT}, e.g. AlphaStar~\cite{Vinyals2019,alpha_star_es}. 

Finally, low-rank tensor approximations have been applied to RL problems in settings different from ours, including multi-agent scenarios ~\cite{mahajan2021tesseract}, and improving dynamic programming approaches~\cite{Gorodetsky2018,boyko2021tt}. The idea of our method is quite different from the presented works, especially in the RL area.

\section{Conclusion}
\label{sec:conc}
We proposed a new discrete optimization method based on quantized tensor-train representation and maximal volume principle. We demonstrate its performance for analytical benchmark functions and reinforcement learning problems.
Our algorithm is more efficient under a fixed computational budget than 
baselines, especially on discrete domains. Moreover, the execution time for TTOpt is lower by a significant margin compared with other baselines.
Finally, we show that the agents with discrete parameters in RL can be as efficient as their continuous parameter versions. This observation supports the broad adoption of quantization in machine learning. We hope that our approach will serve as a bridge between continuous and discrete optimization methods.

\begin{ack}
The work was supported by Ministry of Science and Higher Education grant No. 075-10-2021-068.
\end{ack}

\bibliographystyle{plain}
\bibliography{bibliography}

\begin{thebibliography}{10}

\bibitem{ahmadi2021cross}
Salman Ahmadi-Asl, Cesar~F Caiafa, Andrzej Cichocki, Anh~Huy Phan, Toshihisa
  Tanaka, Ivan Oseledets, and Jun Wang.
\newblock Cross tensor approximation methods for compression and dimensionality
  reduction.
\newblock {\em IEEE Access}, 9:150809--150838, 2021.

\bibitem{ALARIE2021100011}
Stéphane Alarie, Charles Audet, Aïmen~E. Gheribi, Michael Kokkolaras, and
  Sébastien {Le Digabel}.
\newblock Two decades of blackbox optimization applications.
\newblock {\em EURO Journal on Computational Optimization}, 9:100011, 2021.

\bibitem{alpha_star_es}
Kai Arulkumaran, Antoine Cully, and Julian Togelius.
\newblock Alphastar: An evolutionary computation perspective.
\newblock In {\em Proceedings of the Genetic and Evolutionary Computation
  Conference Companion}, GECCO '19, page 314–315, New York, NY, USA, 2019.
  Association for Computing Machinery.

\bibitem{1562233}
A.F. Atiya, A.G. Parlos, and L.~Ingber.
\newblock A reinforcement learning method based on adaptive simulated
  annealing.
\newblock In {\em 2003 46th Midwest Symposium on Circuits and Systems},
  volume~1, pages 121--124 Vol. 1, 2003.

\bibitem{Nevergrad}
Pauline Bennet, Carola Doerr, Antoine Moreau, Jeremy Rapin, Fabien Teytaud, and
  Olivier Teytaud.
\newblock Nevergrad: Black-box optimization platform.
\newblock {\em SIGEVOlution}, 14(1):8–15, apr 2021.

\bibitem{constraint-es2}
Rafał Biedrzycki.
\newblock Handling bound constraints in cma-es: An experimental study.
\newblock {\em Swarm and Evolutionary Computation}, 52:100627, 2020.

\bibitem{boyko2021tt}
AI~Boyko, IV~Oseledets, and G~Ferrer.
\newblock Tt-qi: Faster value iteration in tensor train format for stochastic
  optimal control.
\newblock {\em Computational Mathematics and Mathematical Physics},
  61(5):836--846, 2021.

\bibitem{Brockman2016OpenAIG}
Greg Brockman, Vicki Cheung, Ludwig Pettersson, Jonas Schneider, John Schulman,
  Jie Tang, and Wojciech Zaremba.
\newblock Openai gym.
\newblock {\em ArXiv}, abs/1606.01540, 2016.

\bibitem{CAIAFA2010557}
Cesar~F. Caiafa and Andrzej Cichocki.
\newblock Generalizing the column–row matrix decomposition to multi-way
  arrays.
\newblock {\em Linear Algebra and its Applications}, 433(3):557--573, 2010.

\bibitem{pmlr-v80-choromanski18a}
Krzysztof Choromanski, Mark Rowland, Vikas Sindhwani, Richard Turner, and
  Adrian Weller.
\newblock Structured evolution with compact architectures for scalable policy
  optimization.
\newblock In Jennifer Dy and Andreas Krause, editors, {\em Proceedings of the
  35th International Conference on Machine Learning}, volume~80 of {\em
  Proceedings of Machine Learning Research}, pages 970--978. PMLR, 10--15 Jul
  2018.

\bibitem{NEURIPS2019_asebo}
Krzysztof~M Choromanski, Aldo Pacchiano, Jack Parker-Holder, Yunhao Tang, and
  Vikas Sindhwani.
\newblock From complexity to simplicity: Adaptive es-active subspaces for
  blackbox optimization.
\newblock In H.~Wallach, H.~Larochelle, A.~Beygelzimer, F.~d\textquotesingle
  Alch\'{e}-Buc, E.~Fox, and R.~Garnett, editors, {\em Advances in Neural
  Information Processing Systems}, volume~32. Curran Associates, Inc., 2019.

\bibitem{blind}
Andrzej Cichocki and Shun-ichi Amari.
\newblock {\em Adaptive Blind Signal and Image Processing: Learning Algorithms
  and Applications}.
\newblock John Wiley \& Sons, Inc., USA, 2002.

\bibitem{CichockiBookPart1MAL059}
Andrzej Cichocki, Namgil Lee, Ivan Oseledets, Anh-Huy Phan, Qibin Zhao, and
  Danilo~P. Mandic.
\newblock Tensor networks for dimensionality reduction and large-scale
  optimization: Part 1 low-rank tensor decompositions.
\newblock {\em Foundations and Trends® in Machine Learning}, 9(4-5):249--429,
  2016.

\bibitem{ActiveSubspace}
Paul~G. Constantine.
\newblock Active subspaces - emerging ideas for dimension reduction in
  parameter studies.
\newblock In {\em SIAM spotlights}, 2015.

\bibitem{conti2018}
Edoardo Conti, Vashisht Madhavan, Felipe~Petroski Such, Joel Lehman, Kenneth~O.
  Stanley, and Jeff Clune.
\newblock Improving exploration in evolution strategies for deep reinforcement
  learning via a population of novelty-seeking agents.
\newblock In {\em Proceedings of the 32nd International Conference on Neural
  Information Processing Systems}, NIPS'18, page 5032–5043, Red Hook, NY,
  USA, 2018. Curran Associates Inc.

\bibitem{Coulom-2002a}
R\'emi Coulom.
\newblock {\em Reinforcement Learning Using Neural Networks, with Applications
  to Motor Control}.
\newblock PhD thesis, Institut National Polytechnique de Grenoble, 2002.

\bibitem{ErezTT11}
Tom Erez, Yuval Tassa, and Emanuel Todorov.
\newblock Infinite-horizon model predictive control for periodic tasks with
  contacts.
\newblock In Hugh~F. Durrant{-}Whyte, Nicholas Roy, and Pieter Abbeel, editors,
  {\em Robotics: Science and Systems VII, University of Southern California,
  Los Angeles, CA, USA, June 27-30, 2011}, 2011.

\bibitem{Faust2019EvolvingRT}
Aleksandra Faust, Anthony~G. Francis, and Dar Mehta.
\newblock Evolving rewards to automate reinforcement learning.
\newblock In {\em 6th ICML Workshop on Automated Machine Learning}, 2019.

\bibitem{goreinov2010find}
Sergei~A Goreinov, Ivan~V Oseledets, Dimitry~V Savostyanov, Eugene~E
  Tyrtyshnikov, and Nikolay~L Zamarashkin.
\newblock How to find a good submatrix.
\newblock In {\em Matrix Methods: Theory, Algorithms And Applications:
  Dedicated to the Memory of Gene Golub}, pages 247--256. World Scientific,
  2010.

\bibitem{Gorodetsky2018}
Alex Gorodetsky, Sertac Karaman, and Youssef Marzouk.
\newblock High-dimensional stochastic optimal control using continuous tensor
  decompositions.
\newblock {\em The International Journal of Robotics Research},
  37(2-3):340--377, 2018.

\bibitem{NEURIPS2018_worldmodels}
David Ha and J\"{u}rgen Schmidhuber.
\newblock Recurrent world models facilitate policy evolution.
\newblock In S.~Bengio, H.~Wallach, H.~Larochelle, K.~Grauman, N.~Cesa-Bianchi,
  and R.~Garnett, editors, {\em Advances in Neural Information Processing
  Systems}, volume~31. Curran Associates, Inc., 2018.

\bibitem{hackbusch2009new}
Wolfgang Hackbusch and Stefan K{\"u}hn.
\newblock A new scheme for the tensor representation.
\newblock {\em Journal of Fourier analysis and applications}, 15(5):706--722,
  2009.

\bibitem{Hansen2006}
Nikolaus Hansen.
\newblock {\em The CMA Evolution Strategy: A Comparing Review}, pages 75--102.
\newblock Springer Berlin Heidelberg, Berlin, Heidelberg, 2006.

\bibitem{HEIDRICHMEISNER2009152}
Verena Heidrich-Meisner and Christian Igel.
\newblock Neuroevolution strategies for episodic reinforcement learning.
\newblock {\em Journal of Algorithms}, 64(4):152--168, 2009.
\newblock Special Issue: Reinforcement Learning.

\bibitem{HEIN201787}
Daniel Hein, Alexander Hentschel, Thomas Runkler, and Steffen Udluft.
\newblock Particle swarm optimization for generating interpretable fuzzy
  reinforcement learning policies.
\newblock {\em Engineering Applications of Artificial Intelligence}, 65:87--98,
  2017.

\bibitem{HerrmannPriceJoyce+2020+45+62}
J.~Michael Herrmann, Adam Price, and Thomas Joyce.
\newblock {\em 3. Ant colony optimization and reinforcement learning}, pages
  45--62.
\newblock De Gruyter, 2020.

\bibitem{holland92}
John~H. Holland.
\newblock Genetic algorithms.
\newblock {\em Scientific American}, 267(1):66--73, 1992.

\bibitem{holtz2012alternating}
Sebastian Holtz, Thorsten Rohwedder, and Reinhold Schneider.
\newblock The alternating linear scheme for tensor optimization in the tensor
  train format.
\newblock {\em SIAM Journal on Scientific Computing}, 34(2):A683--A713, 2012.

\bibitem{jamil2013literature}
Momin Jamil and Xin-She Yang.
\newblock A literature survey of benchmark functions for global optimisation
  problems.
\newblock {\em International Journal of Mathematical Modelling and Numerical
  Optimisation}, 4(2):150--194, 2013.

\bibitem{10.5555/3326943.3327053}
Shauharda Khadka and Kagan Tumer.
\newblock Evolution-guided policy gradient in reinforcement learning.
\newblock In {\em Proceedings of the 32nd International Conference on Neural
  Information Processing Systems}, NIPS'18, page 1196–1208, Red Hook, NY,
  USA, 2018. Curran Associates Inc.

\bibitem{Khoromskij2011}
Boris~N. Khoromskij.
\newblock O(dlog{\thinspace}n)-quantics approximation of n-d tensors in
  high-dimensional numerical modeling.
\newblock {\em Constructive Approximation}, 34(2):257--280, Oct 2011.

\bibitem{kolda2003ds}
Tamara~G. Kolda, Robert~Michael Lewis, and Virginia Torczon.
\newblock Optimization by direct search: New perspectives on some classical and
  modern methods.
\newblock {\em SIAM Review}, 45(3):385--482, 2003.

\bibitem{constraint-es1}
Oliver Kramer.
\newblock A review of constraint-handling techniques for evolution strategies.
\newblock {\em Appl. Comp. Intell. Soft Comput.}, 2010, January 2010.

\bibitem{es-not-fd-approx}
Joel Lehman, Jay Chen, Jeff Clune, and Kenneth~O. Stanley.
\newblock Es is more than just a traditional finite-difference approximator.
\newblock In {\em Proceedings of the Genetic and Evolutionary Computation
  Conference}, GECCO '18, page 450–457, New York, NY, USA, 2018. Association
  for Computing Machinery.

\bibitem{self-guided-es}
Fei-Yu Liu, Zi-Niu Li, and Chao Qian.
\newblock Self-guided evolution strategies with historical estimated gradients.
\newblock In {\em Proceedings of the Twenty-Ninth International Joint
  Conference on Artificial Intelligence}, IJCAI'20, 2021.

\bibitem{10.1016/j.engappai.2020.103525}
Tundong Liu, Liduan Li, Guifang Shao, Xiaomin Wu, and Meng Huang.
\newblock A novel policy gradient algorithm with pso-based parameter
  exploration for continuous control.
\newblock {\em Eng. Appl. Artif. Intell.}, 90(C), apr 2020.

\bibitem{mahajan2021tesseract}
Anuj Mahajan, Mikayel Samvelyan, Lei Mao, Viktor Makoviychuk, Animesh Garg,
  Jean Kossaifi, Shimon Whiteson, Yuke Zhu, and Animashree Anandkumar.
\newblock Tesseract: Tensorised actors for multi-agent reinforcement learning.
\newblock In {\em International Conference on Machine Learning (ICML)}, volume
  139, pages 7301--7312, 2021.

\bibitem{pmlr-v97-maheswaranathan19a}
Niru Maheswaranathan, Luke Metz, George Tucker, Dami Choi, and Jascha
  Sohl-Dickstein.
\newblock Guided evolutionary strategies: augmenting random search with
  surrogate gradients.
\newblock In Kamalika Chaudhuri and Ruslan Salakhutdinov, editors, {\em
  Proceedings of the 36th International Conference on Machine Learning},
  volume~97 of {\em Proceedings of Machine Learning Research}, pages
  4264--4273. PMLR, 09--15 Jun 2019.

\bibitem{NEURIPS2018_ARS}
Horia Mania, Aurelia Guy, and Benjamin Recht.
\newblock Simple random search of static linear policies is competitive for
  reinforcement learning.
\newblock In S.~Bengio, H.~Wallach, H.~Larochelle, K.~Grauman, N.~Cesa-Bianchi,
  and R.~Garnett, editors, {\em Advances in Neural Information Processing
  Systems}, volume~31. Curran Associates, Inc., 2018.

\bibitem{JMLR:v22:18-220}
Erich Merrill, Alan Fern, Xiaoli Fern, and Nima Dolatnia.
\newblock An empirical study of bayesian optimization: Acquisition versus
  partition.
\newblock {\em Journal of Machine Learning Research}, 22(4):1--25, 2021.

\bibitem{9524335}
Laurent Meunier, Herilalaina Rakotoarison, Pak~Kan Wong, Baptiste Roziere,
  Jérémy Rapin, Olivier Teytaud, Antoine Moreau, and Carola Doerr.
\newblock Black-box optimization revisited: Improving algorithm selection
  wizards through massive benchmarking.
\newblock {\em IEEE Transactions on Evolutionary Computation}, pages 1--1,
  2021.

\bibitem{mikhalev2018rectangular}
Aleksandr Mikhalev and Ivan~V Oseledets.
\newblock Rectangular maximum-volume submatrices and their applications.
\newblock {\em Linear Algebra and its Applications}, 538:187--211, 2018.

\bibitem{Mnih2015}
Volodymyr Mnih, Koray Kavukcuoglu, David Silver, Andrei~A. Rusu, Joel Veness,
  Marc~G. Bellemare, Alex Graves, Martin Riedmiller, Andreas~K. Fidjeland,
  Georg Ostrovski, Stig Petersen, Charles Beattie, Amir Sadik, Ioannis
  Antonoglou, Helen King, Dharshan Kumaran, Daan Wierstra, Shane Legg, and
  Demis Hassabis.
\newblock Human-level control through deep reinforcement learning.
\newblock {\em Nature}, 518(7540):529--533, Feb 2015.

\bibitem{NeldMead65}
John~A. Nelder and Roger Mead.
\newblock A simplex method for function minimization.
\newblock {\em Computer Journal}, 7:308--313, 1965.

\bibitem{Nesterov2017}
Yurii Nesterov and Vladimir Spokoiny.
\newblock Random gradient-free minimization of convex functions.
\newblock {\em Foundations of Computational Mathematics}, 17(2):527--566, Apr
  2017.

\bibitem{7379496}
Barry~D. Nichols.
\newblock Continuous action-space reinforcement learning methods applied to the
  minimum-time swing-up of the acrobot.
\newblock In {\em 2015 IEEE International Conference on Systems, Man, and
  Cybernetics}, pages 2084--2089, 2015.

\bibitem{oseledets2010approximation}
I.~V. Oseledets.
\newblock Approximation of $2^d \times 2^d$ matrices using tensor
  decomposition.
\newblock {\em SIAM J. Matrix Anal. Appl.}, 31(4):2130--2145, 2010.

\bibitem{oseledets2011tensor}
I.~V. Oseledets.
\newblock Tensor-train decomposition.
\newblock {\em SIAM Journal on Scientific Computing}, 33(5):2295--2317, 2011.

\bibitem{oseledets2009breaking}
Ivan~V Oseledets and Eugene~E Tyrtyshnikov.
\newblock Breaking the curse of dimensionality, or how to use svd in many
  dimensions.
\newblock {\em SIAM Journal on Scientific Computing}, 31(5):3744--3759, 2009.

\bibitem{oseledets2010ttcross}
Ivan~V Oseledets and Eugene~E Tyrtyshnikov.
\newblock {TT-cross} approximation for multidimensional arrays.
\newblock {\em Linear Algebra and its Applications}, 432(1):70--88, 2010.

\bibitem{PhysRevResearch.2.033429}
David Pfau, James~S. Spencer, Alexander G. D.~G. Matthews, and W.~M.~C.
  Foulkes.
\newblock Ab initio solution of the many-electron schr\"odinger equation with
  deep neural networks.
\newblock {\em Phys. Rev. Research}, 2:033429, Sep 2020.

\bibitem{phan2020a}
Anh-Huy Phan, Andrzej Cichocki, André Uschmajew, Petr Tichavský, George Luta,
  and Danilo~P. Mandic.
\newblock Tensor networks for latent variable analysis: Novel algorithms for
  tensor train approximation.
\newblock {\em IEEE Transactions on Neural Networks and Learning Systems},
  31(11):4622--4636, 2020.

\bibitem{pmlr-v32-qin14}
Zhiwei Qin, Weichang Li, and Firdaus Janoos.
\newblock Sparse reinforcement learning via convex optimization.
\newblock In Eric~P. Xing and Tony Jebara, editors, {\em Proceedings of the
  31st International Conference on Machine Learning}, volume~32 of {\em
  Proceedings of Machine Learning Research}, pages 424--432, Bejing, China,
  22--24 Jun 2014. PMLR.

\bibitem{doi:10.1177/027836498400300207}
Marc~H. Raibert, Jr~H.~Benjamin~Brown, and Michael Chepponis.
\newblock Experiments in balance with a 3d one-legged hopping machine.
\newblock {\em The International Journal of Robotics Research}, 3(2):75--92,
  1984.

\bibitem{Ramesh2021ZeroShotTG}
Aditya Ramesh, Mikhail Pavlov, Gabriel Goh, Scott Gray, Chelsea Voss, Alec
  Radford, Mark Chen, and Ilya Sutskever.
\newblock Zero-shot text-to-image generation.
\newblock {\em ArXiv}, abs/2102.12092, 2021.

\bibitem{Salimans2017EvolutionSA}
Tim Salimans, Jonathan Ho, Xi~Chen, and Ilya Sutskever.
\newblock Evolution strategies as a scalable alternative to reinforcement
  learning.
\newblock {\em ArXiv}, abs/1703.03864, 2017.

\bibitem{trpo-pmlr-v37-schulman15}
John Schulman, Sergey Levine, Pieter Abbeel, Michael Jordan, and Philipp
  Moritz.
\newblock Trust region policy optimization.
\newblock In Francis Bach and David Blei, editors, {\em Proceedings of the 32nd
  International Conference on Machine Learning}, volume~37 of {\em Proceedings
  of Machine Learning Research}, pages 1889--1897, Lille, France, 07--09 Jul
  2015. PMLR.

\bibitem{ppo-Schulman2017ProximalPO}
John Schulman, Filip Wolski, Prafulla Dhariwal, Alec Radford, and Oleg Klimov.
\newblock Proximal policy optimization algorithms.
\newblock {\em ArXiv}, abs/1707.06347, 2017.

\bibitem{Schwefel1977}
Hans-Paul Schwefel.
\newblock {\em Evolutionsstrategien f{\"u}r die numerische Optimierung}, pages
  123--176.
\newblock Birkh{\"a}user Basel, Basel, 1977.

\bibitem{Schweidtmann_2018}
Artur~M. Schweidtmann and Alexander Mitsos.
\newblock Deterministic global optimization with artificial neural networks
  embedded.
\newblock {\em Journal of Optimization Theory and Applications},
  180(3):925–948, Oct 2018.

\bibitem{Storn1997-DE}
Rainer Storn and Kenneth Price.
\newblock Differential evolution -- a simple and efficient heuristic for global
  optimization over continuous spaces.
\newblock {\em Journal of Global Optimization}, 11(4):341--359, Dec 1997.

\bibitem{SimpleGASuch2017-DeepNG}
Felipe~Petroski Such, Vashisht Madhavan, Edoardo Conti, Joel Lehman, Kenneth~O.
  Stanley, and Jeff Clune.
\newblock Deep neuroevolution: Genetic algorithms are a competitive alternative
  for training deep neural networks for reinforcement learning.
\newblock {\em ArXiv}, abs/1712.06567, 2017.

\bibitem{mujoco}
Emanuel Todorov, Tom Erez, and Yuval Tassa.
\newblock Mujoco: A physics engine for model-based control.
\newblock In {\em 2012 IEEE/RSJ International Conference on Intelligent Robots
  and Systems}, pages 5026--5033, 2012.

\bibitem{Vinyals2019}
Oriol Vinyals, Igor Babuschkin, Wojciech~M. Czarnecki, Micha{\"e}l Mathieu,
  Andrew Dudzik, Junyoung Chung, David~H. Choi, Richard Powell, Timo Ewalds,
  Petko Georgiev, Junhyuk Oh, Dan Horgan, Manuel Kroiss, Ivo Danihelka, Aja
  Huang, Laurent Sifre, Trevor Cai, John~P. Agapiou, Max Jaderberg,
  Alexander~S. Vezhnevets, R{\'e}mi Leblond, Tobias Pohlen, Valentin Dalibard,
  David Budden, Yury Sulsky, James Molloy, Tom~L. Paine, Caglar Gulcehre, Ziyu
  Wang, Tobias Pfaff, Yuhuai Wu, Roman Ring, Dani Yogatama, Dario W{\"u}nsch,
  Katrina McKinney, Oliver Smith, Tom Schaul, Timothy Lillicrap, Koray
  Kavukcuoglu, Demis Hassabis, Chris Apps, and David Silver.
\newblock Grandmaster level in starcraft ii using multi-agent reinforcement
  learning.
\newblock {\em Nature}, 575(7782):350--354, Nov 2019.

\bibitem{4400335}
Pawel Wawrzynski.
\newblock Learning to control a 6-degree-of-freedom walking robot.
\newblock In {\em EUROCON 2007 - The International Conference on "Computer as a
  Tool"}, pages 698--705, 2007.

\bibitem{JMLR:v15:wierstra14a}
Daan Wierstra, Tom Schaul, Tobias Glasmachers, Yi~Sun, Jan Peters, and
  J\"{u}rgen Schmidhuber.
\newblock Natural evolution strategies.
\newblock {\em Journal of Machine Learning Research}, 15(27):949--980, 2014.

\bibitem{8682231}
Qibin Zhao, Masashi Sugiyama, Longhao Yuan, and Andrzej Cichocki.
\newblock Learning efficient tensor representations with ring-structured
  networks.
\newblock In {\em ICASSP 2019 - 2019 IEEE International Conference on
  Acoustics, Speech and Signal Processing (ICASSP)}, pages 8608--8612, 2019.

\end{thebibliography}

\section*{Checklist}
\begin{enumerate}

\item For all authors...
\begin{enumerate}
  \item Do the main claims made in the abstract and introduction accurately reflect the paper's contributions and scope?
    \answerYes{}
  \item Did you describe the limitations of your work?
    \answerYes{}
  \item Did you discuss any potential negative societal impacts of your work?
    \answerNA{} The work has no any societal impacts, since its about optimization methods.
  \item Have you read the ethics review guidelines and ensured that your paper conforms to them?
    \answerYes{}
\end{enumerate}

\item If you are including theoretical results...
\begin{enumerate}
  \item Did you state the full set of assumptions of all theoretical results?
    \answerYes{}
        \item Did you include complete proofs of all theoretical results?
    \answerNA{}
\end{enumerate}

\item If you ran experiments...
\begin{enumerate}
  \item Did you include the code, data, and instructions needed to reproduce the main experimental results (either in the supplemental material or as a URL)?
    \answerYes{The code will be provided in supplemental material and as  github repository after decision } 
  \item Did you specify all the training details (e.g., data splits, hyperparameters, how they were chosen)? 
    \answerYes{} 
        \item Did you report error bars (e.g., with respect to the random seed after running experiments multiple times)?
    \answerYes{} We used a typical shade(mean-std) plots that used in   reinforcement learning experiments 
        \item Did you include the total amount of compute and the type of resources used (e.g., type of GPUs, internal cluster, or cloud provider)?
    \answerYes{}
\end{enumerate}

\item If you are using existing assets (e.g., code, data, models) or curating/releasing new assets...
\begin{enumerate}
  \item If your work uses existing assets, did you cite the creators?
    \answerYes{}. We mentioned all references and source-codes for gradient-free baselines.
  \item Did you mention the license of the assets?
    \answerNA{}
  \item Did you include any new assets either in the supplemental material or as a URL?
    \answerNA{}
  \item Did you discuss whether and how consent was obtained from people whose data you're using/curating?
    \answerNA{}
  \item Did you discuss whether the data you are using/curating contains personally identifiable information or offensive content?
    \answerNA{}
\end{enumerate}

\item If you used crowdsourcing or conducted research with human subjects...
\begin{enumerate}
  \item Did you include the full text of instructions given to participants and screenshots, if applicable?
    \answerNA{}
  \item Did you describe any potential participant risks, with links to Institutional Review Board (IRB) approvals, if applicable?
    \answerNA{}
  \item Did you include the estimated hourly wage paid to participants and the total amount spent on participant compensation?
    \answerNA{}
\end{enumerate}

\end{enumerate}


\maketitle
\appendix

\section{Description of MaxVol and TTOpt algorithms}

\subsection{Maxvol Algoritm\label{apx:maxvol}}
\label{sec:appendix_maxvol}

The search for the maximal element in a large matrix can be significantly simplified if one can obtain a "good" submatrix, for example, a maximal-volume submatrix. Finding the maximal-volume submatrix of a nondegenerate matrix $\matr{A} \in \set{R}^{N \times R}, N > R$ is an NP-hard problem. 
In this paper, we adapted the \func{maxvol} algorithm~\cite{goreinov2010find} which can find a  submatrix $\matr{C} \in \set{R}^{R \times R}$ of $\matr{A}$, such that its determinant is close to maximum in absolute value. The algorithm selects a set of $R$ rows denoted by $\vect{I} \subset [1,\ldots, N]$ which form the matrix $\matr{C} = \matr{A}[\vect{I}, :]$.


For the given tolerance threshold $\epsilon$ ($\epsilon \geq 1$ and close to one), we find the set $\vect{I}$ as follows:
\begin{enumerate}
    \item Compute the LU-decomposition $\matr{A} = \matr{P} \matr{L} \matr{U}$ and store the permutation of the first $R$ rows (according to the matrix $\matr{P}$) in the list $I$.
    \item Generate the matrix $\matr{Q} \in \set{R}^{R \times N}$ as a solution to the linear system with an upper triangular matrix $\matr{U}$: $\matr{U}^T \matr{Q} = \matr{A}^T$.
    \item Compute the matrix $\matr{B} \in \set{R}^{N \times R}$ as a solution to the linear system with the lower triangular matrix $\matr{L}$: $(\matr{L}[:R, :])^T \matr{B}^T = \matr{Q}$.
    \item Find the maximum modulo element $b = \matr{B}_{i, j}$ of the matrix $\matr{B}$. If $|b| \leq \epsilon$, then terminate the algorithm\footnote{
        In addition to the indices of the rows $\vect{I}$ that form the maximal-volume submatrix $\matr{C} \in \set{R}^{R \times R}$, we also obtain the matrix of coefficients $\matr{B} \in \set{R}^{N \times R}$ such that $\matr{A} = \matr{B} \matr{C}$.
    } by returning the current list $I$ and matrix $\matr{B}$.
    \item Update the matrix $\matr{B} \gets \matr{B} - \matr{B}[:, j] \, (\matr{B}[i,:] - \vect{e}_j^T) \, b^{-1}$, where $\vect{e}_j$ is the $j$-th unit basis vector.
    \item Update the list $\vect{I}$ as $\vect{I}[j] \gets i$.
    \item Return to step 4.
\end{enumerate}

It can be shown that at each step $k$ of the algorithm the volume of the submatrix $\matr{C}$ increases by a factor not less than $\epsilon$. Therefore, the estimate for the number of iterations, $K$, is the following:
\begin{align}
K =
\frac{
    \log{( \, | \det{(\hat{\matr{C}})} | \, )} -
    \log{( \, | \det{(\matr{C}^{(0)})} | \, )}
}
{
    \log{\epsilon}
},
\end{align}
where $\hat{\matr{C}}$ is the exact maximal-volume submatrix and $\matr{C}^{(0)}$ is an initial approximation obtained from the LU-decomposition (see step 1 above).
The computational complexity of the algorithm is the sum of the initialization complexity and the complexities of $K$ iterations. The complexity equals
\begin{equation}
\order{N R^2 + K N R}.
\end{equation}

In the context of the maximal matrix element search problem, other definitions of "good" submatrices are possible. For example, one can search for a rectangular submatrix containing rows or columns which span the largest volume. The \func{rect\_maxvol} algorithm can be used in this case.
We employ \func{rect\_maxvol}~\cite{mikhalev2018rectangular} to choose rectangular submatrices within the framework of the TTOpt algorithm to adaptively increase the rank.

Consider an arbitrary nondegenerate matrix $\matr{A} \in \set{R}^{N \times R}$ ($N > R$). We search for a rectangular submatrix $\hat{\matr{C}} \in \set{R}^{(R+\Delta R) \times R}, R+\Delta R < N$, which maximizes the rectangular volume $\sqrt{ \, \det{(\hat{\matr{C}}^T \hat{\matr{C})}} \, }$ objective. 
An approximation $\matr{C}$ to $\hat{\matr{C}}$ can be found as follows. The first $R$ rows of $\matr{C}$ are obtained by the \func{maxvol} algorithm. The following \func{rect\_maxvol} algorithm will find additional $\Delta R$ rows:
\begin{enumerate}
    \item First, generate $R$ indices of rows $\vect{I} \in \set{N}^{R}$ and the coefficient matrix $\matr{B} \in \set{R}^{N \times R}$ of the initial approximation using the \func{maxvol} algorithm.
    \item Compute the vector $\vect{l} \in \set{R}^{N}$ containing the norms of the rows: $\vect{l}[j] = (\matr{B}[j, :])^T \matr{B}[j, :]$  for $j = 1, 2, \ldots, N$.
    \item Find the maximum modulo element of the vector $\vect{l}$, i.e. $i = argmax(\vect{l})$.
    \item If the current length $R + \Delta R$ of the vector $\vect{I}$ is greater than or equal to $R^{(max)}$ or if $\vect{l}[i] \leq \tau^2$, then terminate the algorithm\footnote{
        As a criterion for stopping the algorithm, we consider either the achievement of the maximum number of rows in the maximal-volume submatrix ($R^{(max)}$), or the sufficiently small norm of the remaining rows in the matrix of coefficients $\matr{B}$.
    } by returning the current list $\vect{I}$ and matrix $\matr{B}$.
    \item Update the coefficient matrix as
        $$
        {\matr{B}} := \begin{bmatrix}
            {\matr{B}}
            -
            \frac
            {
                \matr{B} (\matr{B}[i,:])^T \matr{B}[i,:]
            }
            {
                1 + \matr{B}[i,:] (\matr{B}[i,:])^T
            }
            &
            \frac
            {
                \matr{B} (\matr{B}[i,:])^{T}
            }
            {
                1 + \matr{B}[i,:] (\matr{B}[i,:])^T
            }
        \end{bmatrix}.
        $$
    \item Update the vector of row norms
        $$
        \vect{l}[j] :=
        \vect{l}[j]
        -
        \frac{
            |\matr{B}[j,:] (\matr{B}[i,:])^T|^2
        }
        {
            1 + \matr{B}[i,:] (\matr{B}[i,:])^T
        },
        \quad
        j = 1, 2, \ldots, N.
        $$
    \item Add current row index $i$ to the list $\vect{I}$.
    \item Return to step 3.
\end{enumerate}

The approximate computational complexity of this algorithm according to the work~\cite{mikhalev2018rectangular} is $O(N R^2)$, and the expected number of rows in the resulting submatrix is $2 R - 1$ for the case $\tau = 1$.

\subsection{TTOpt Algorithm}
\label{sec:appendix_ttopt}
In Algorithm~\ref{alg:ttmin} we present the details of the \func{TTOpt} implementation. The \func{TTOpt} algorithm builds the TT proxy of the minimized function which is iteratively updated. 
The procedure of updating the TT proxy is outlined in Algorithm~\ref{alg:ttmin_update_left} (function \func{update\_left}, that  updates core tensors of the network from right to left ) and in Algorithm~\ref{alg:ttmin_update_right} (function \func{update\_right}, updates core tensors of the network from left to right ).
The requests to the objective function and the transformation of resulting values are presented in Algorithm~\ref{alg:ttmin_eval} (function \func{eval}).

The procedure begins by building one-dimensional uniform grids $\vx_i$ ($i = 1, 2, \ldots, d$) for the function argument along each mode, using the specified bounds of the rectangular search domain $\Omega$.
Note that, if necessary, arbitrary nonuniform grids can be used, taking into account the specific features of the function under consideration.

\begin{figure}[H]
\vskip -0.1in
\begin{center}
\begin{algorithm}[H]
\small
\SetAlgoLined
\KwData{
    function $\fj(\vt)$, where $\vt \in \Omega \subset \set{R}^d$;
    boundary points of the rectangular domain $\Omega = [a_1, b_1] \times [a_2, b_2] \times \cdots \times [a_d, b_d]$;
    the number of grid points for every dimension $N_1, N_2, \ldots, N_d$;
    number of inner iterations (sweeps) $k_{\textit{max}}$;
    maximum TT-rank $r_{\textit{max}}$.
}
\KwResult{
    approximation of the spatial point $\vt_{min} \in \set{R}^d$ at which the function $\fj$ reaches its minimum in the region $\Omega$ and a corresponding function value $\fj_{min} \in \set{R}$.
}

// Construct a uniform grid $\vx_1, \vx_2, \ldots, \vx_d$:

Set $
    \vx_i [m] = a_i + (b_i - a_i) \cdot \frac{m - 1}{N_i - 1}
$

for all $i = 1, \ldots, d$ and $m = 1, 2, \ldots, N_i$.

// Initialize the TT-ranks $R_0, R_1, \ldots, R_d$:

Set $R_0 = 1$.

\For{$i = 1$ \KwTo $(d-1)$}{
    Set $R_i = \min{(R_{i-1} \cdot N_{i}, R_i \cdot N_i, r_{\textit{max}})}$.
}

Set $R_d = 1$.     

// Initialize set of points of interest $\matr{X}_0, \matr{X}_1, \ldots, \matr{X}_d$:

Set $\matr{X}_{0} = None$. 

\For{$i = 0$ \KwTo $(d-2)$}{
    Set $
        \matr{G}_i = \func{random}(R_i \cdot N_i, R_{i+1})
    $ from the standard normal distribution.
    
    Compute QR-decomposition $
        \matr{Q}, \matr{R} = \func{QR}(\matr{G}_i)
    $.
    
    Compute indices $
        \vect{m}_{opt} = \func{maxvol}(\matr{Q})
    $.
    
    Set $
        \matr{X}_{i+1}
        =
        \func{update\_right}
            (\matr{X}_{i}, \vx_i, N_i, R_i, \vect{m}_{opt})
    $.
}

Set $\matr{X}_d = None$. 

// Iterate in a loop to find optimal $\vt_{min}$ and $\fj_{min}$:

Set
$
    \vt_{min} = None
$
and
$
    \fj_{min} = + \infty
$.

\For{$k = 1$ \KwTo $k_{\textit{max}}$}{
    // Traverse the TT-cores from right to left:
    
    \For{$i = d$ \KwTo $1$}{
        Compute $
            \vz, \vt_{min}, \fj_{min} =
                \func{eval}(*)
        $.
        
        Reshape $\vz$ to matrix
        $
            \mz \in \set{R}^{R_i \times n_i \cdot R_{i+1}}
        $.
        
        Compute
        $
            \matr{Q}, \matr{R} = \func{QR}(\mz^T)
        $.

        Compute indices $
            \vect{m}_{opt} = \func{rect\_maxvol}(\matr{Q}[:, 0:R[i]])
        $.
        
        Set $
            \matr{X}_{i}
            =
            \func{update\_left}
                (\matr{X}_{i+1}, \vx_i, N_i, R_{i+1}, \vect{m}_{opt})
        $.
    }

    // Traverse the TT-cores from left to right:
    
    \For{$i = 1$ \KwTo $d$}{
        Compute $
            \vz, \vt_{min}, \fj_{min} = \func{eval}(*)
        $.
        
        Reshape $\vz$ to matrix
        $
            \vz \in \set{R}^{R_i \cdot n_i \times R_{i+1}}
        $.

        Compute
        $
            \matr{Q}, \matr{R} = \func{QR}(\mz)
        $.

        Compute indices $
            \vect{m}_{opt} = \func{rect\_maxvol}(\matr{Q})
        $.
        
        Set $
            \matr{X}_{i}
            =
            \func{update\_right}
                (\matr{X}_{i+1}, \vx_i, N_i, R_{i+1}, \vect{m}_{opt})
        $.
    }
}

\Return{($\vt_{min}$, $\fj_{min}$)}.

\caption{Multivariable function minimizer \func{TTOpt}.}

\label{alg:ttmin}

\end{algorithm}
\end{center}
\end{figure}

\begin{figure}[H]
\vskip -0.1in
\begin{center}
\begin{algorithm}[H]
\SetAlgoLined
\KwData{
    current set of points for the $(i+1)$-th mode $\matr{X}_{i+1}$; 
    grid points $\vx_i$;
    number of grid points $N_i$;
    TT-rank $R_{i+1}$;
    list of indices to be selected $\vect{m}_{opt}$.
}
\KwResult{
    new set of points $\matr{X}_{i}$.
}

Set
$\matr{W}_1 = \func{ones}(R_{i+1}) \otimes \vect{x}_{i}$
and
$\matr{W}_2 = \matr{X}_{i+1} \otimes \func{ones}(N_i)$

Set $\matr{X}_i = [\matr{W}_1, \matr{W}_2]$.


Select subset of rows $\matr{X}_i = \matr{X}_i [\vect{m}_{opt}, :]$.

\Return{$\matr{X}_{i}$}.

\caption{Function \func{update\_left} to update points of interest when traverse the tensor modes from right to left.}

\label{alg:ttmin_update_left}

\end{algorithm}
\end{center}
\end{figure}

\begin{figure}[H]
\vskip -0.1in
\begin{center}
\begin{algorithm}[H]

\SetAlgoLined
\KwData{
    current set of points for the $i$-th mode $\matr{X}_{i}$; 
    grid points $\vx_i$;
    number of grid points $N_i$;
    TT-rank $R_i$;
    list of indices to be selected $\vect{m}_{opt}$.
}
\KwResult{
    new set of points $\matr{X}_{i+1}$.
}

Set
$\matr{W}_1 = \func{ones}(N_i) \otimes \matr{X}_{i}$
and
$\matr{W}_2 = \vect{x}_{i} \otimes \func{ones}(R_i)$.

Set $\matr{X}_{i+1} = [\matr{W}_1, \matr{W}_2]$.


Select subset of rows $\matr{X}_{i+1} = \matr{X}_{i+1} [\vect{m}_{opt}, :]$.

\Return{$\matr{X}_{i+1}$}.

\caption{Function \func{update\_right} to update points of interest when traverse the tensor modes from left to right.}

\label{alg:ttmin_update_right}

\end{algorithm}
\end{center}
\end{figure}

\begin{figure}[H]
\vskip -0.1in
\begin{center}
\begin{algorithm}[H]
\SetAlgoLined
\KwData{
    $\mx_i$;
    $\mx_{i+1}$;
    $R_i$;
    $N_i$;
    $R_{i+1}$;
    $\fj$;
    $\vt_{min}$,
    $\fj_{min}$
    (see Algorithm~\ref{alg:ttmin} for details).
}
\KwResult{
    transformed function values $\vz \in \set{R}^{R_i \cdot N_i \cdot R_{i+1}}$,
    updated $\vt_{min}$ and $\fj_{min}$.
}

Set $
    \mw_1 =
        \func{ones}(N_i \cdot R_{i+1}) \otimes \mx_i
$.

Set $
    \mw_2 =
        \func{ones}(R_{i+1}) \otimes
        \vx_i \otimes
        \func{ones}(R_i)
$.

Set $
    \mw_3 =
        \mx_{i+1} \otimes \func{ones}(R_i \cdot N_i)
$.

Set $
    \mx^{curr} =
        [\mw_1, \mw_2, \mw_3]
    \in \set{R}^{R_i \cdot N_i \cdot R_{i+1} \times d}$.

// Compute function for each point in $\mx^{curr}$:

Set $\vy^{curr} = \fj(\mx^{curr})$.

\If{$\min{(\vy^{curr})} < \fj_{min}$}{
    Set
    $
        m_{min} = argmin(\vy^{curr})
    $.
    
    Set
    $
        \vt_{min} = \mx^{curr}[m_{min}, :]
    $.
    
    Set
    $
        \fj_{min} = \vy^{curr}[m_{min}]
    $.
}

// Compute smooth function for each value in $\vy^{curr}$:

Set
$
    \vz = \frac{\pi}{2} - \arctan{(\vy^{curr} - \fj_{min})}
$.

\Return{$(\vz, \vt_{min}, \fj_{min})$}.

\caption{Function \func{eval} to compute the target function in points of interest.}

\label{alg:ttmin_eval}

\end{algorithm}
\end{center}
\end{figure}

We then randomly initialize the TT proxy tensor with input ranks $r_{\textit{max}}$. If necessary, we reduce some of the ranks to satisfy the condition $R_{i-1} N_{i} \geq R_i$ ($i = 1, 2, \ldots, d-1$).
Note that in this case, for all TT-cores $\tens{G}_{i} \in \set{R}^{R_{i-1} \times N_i \times R_i}$ ($i = 1, 2, \ldots, d$), the right unfolding matrices $\matr{G}_{i}^{(2)} \in \set{R}^{R_{i-1} \cdot N_i \times R_i}$ will turn out to be ``tall'' matrices, that is, their number of rows is not less than the number of columns, and hence we can apply  $\func{maxvol}$ and $\func{rect\_maxvol}$ algorithms to these matrices.


Next, we iteratively traverse all tensor modes (using corresponding TT-cores) in the direction from right to left and vice versa. We evaluate (and transform) the objective function to refine the selected rows and columns.
For each $k$-th mode of the tensor we evaluate the submatrix $\matr{J}_{k}^{(C)} \in \set{R}^{R_{k-1} \cdot N_k \times R_k}$ of the corresponding unfolding matrix, compute its QR decomposition, find the row indices of the rectangular maximal-volume submatrix $\hat{\matr{J}}_k \in \set{R}^{(R_k + \Delta R_k) \times R_k}$ of the \matr{Q} factor and add resulting indices of the original tensor to the index set $\matr{X}_k$.

The arguments for target function evaluation in Algorithm~\ref{alg:ttmin_eval} are selected as merged left and right index sets, constructed from previous \func{rect\_maxvol} computations.
After each request to the objective function, we update the current optimal value $J_{min}$ and then transform the calculated values by the mapping~(7) described in the main text.

\section{Additional Experiments}
\label{sec:appendix_rles}

\subsection{Experiments with benchmark functions}
\label{sec:appendix_benchmark}

In Section~3.1
we compared the \func{TTOpt} solver\footnote{
    We implemented the \func{TTOpt} algorithm in a python package with detailed documentation, demos, and reproducible scripts for all benchmark calculations.
} with baseline methods, applied to various model functions~\cite{jamil2013literature}.
The list of functions is presented in Table~\ref{tab:compare_benchmark_list}.
For each function, we provide the lower/upper grid bounds ($a$ and $b$) and global minimum ($J_{min}$).
Note that many benchmarks
is multimodal (have two or more local optima), introducing additional complications into the optimization problem.

The main configurable parameters of our solver are the mode size ($N$; $N = 2^q$ in the case of quantization, where $q$ is the number of submodes in the quantized tensor); the rank ($R$), and the limit on the number of requests to the objective function ($M$).
The choice of these parameters can affect the final accuracy of the optimization process.
Below we present the results of the studies of parameter importance.
In all calculations, we fixed the non-varying parameters at the values provided in Section~3.1.

\begin{table}[t!]
\caption{
    Benchmark functions for comparison of the considered optimization algorithms and performance evaluation of the \func{TTOpt} approach.
    For each function, we present the lower grid bound ($a$), the upper grid bound ($b$), the global minimum ($J_{min}$) and the analytical formula.
    Note that $J_{min}$ for the \emph{F6} function is given for the 10-dimensional case.}
\centering

\renewcommand{\arraystretch}{2.2}

\begin{tabular}{|p{2.0cm}|p{1.1cm}|p{1.1cm}|p{1.3cm}|p{6.4cm}|}\hline

Function  &
$a$       &
$b$       &
$J_{min}$ &
Formula   \\ \hline

\emph{F1} \newline (Ackley) &
$-32.768$ &
$32.768$  &
$0.$      &
$
\func{f}(\vx)
=
- A e^{-B \sqrt{
    \frac{1}{d} \sum_{i=1}^d x_i^2
}}
- e^{
    \frac{1}{d} \sum_{i=1}^d \cos{(C x_i)}
}
+ A
+ e^{1},
$
where $A = 20$, $B = 0.2$ and $C = 2 \pi$
\\ \hline

\emph{F2} \newline (Alpine) &
$-10$ &
$10$  &
$0.$  &
$
\func{f}(\vx)
=
\sum_{i=1}^{d}
    | x_i \sin{x_i} + 0.1 x_i |
$
\\ \hline

\emph{F3} \newline (Brown) &
$-1$ &
$4$  &
$0.$ &
$
\func{f}(\vx)
=
\sum_{i=1}^{d-1}
    \left( x_i^2    \right)^{(x_{i+1}^2 + 1)} +
    \left(x_{i+1}^2 \right)^{(x_{i}^2   + 1)}
$
\\ \hline

\emph{F4} \newline (Exponential) &
$-1$  &
$1$   &
$-1.$ &
$
\func{f}(\vx)
= - e^{
    - \frac{1}{2}
    \sum_{i=1}^{d} x_i^2
}
$
\\ \hline

\emph{F5} \newline (Griewank) &
$-600$ &
$600$  &
$0.$   &
$
\func{f}(\vx)
=
\sum_{i=1}^d \frac{x_i^2}{4000} -
\prod_{i=1}^d \cos{\left(\frac{x_i}{\sqrt{i}}\right)} + 1
$
\\ \hline

\emph{F6} \newline (Michalewicz) &
$0$        &
$\pi$      &
$-9.66015$ &
$
\func{f}(\vx)
=
-\sum_{i=1}^d \sin{\left( x_i \right)} \sin^{2m}{\left( \frac{i x_i^2}{\pi} \right)}
$
\\ \hline

\emph{F7} \newline (Qing) &
$0$   &
$500$ &
$0.$  &
$
\func{f}(\vx)
= \sum_{i=1}^d
    \left( x_i^2 - i \right)^2
$
\\ \hline

\emph{F8} \newline (Rastrigin) &
$-5.12$ &
$5.12$  &
$0.$    &
$
\func{f}(\vx)
=
A \cdot d + \sum_{i=1}^{d} \left(
    x_i^2 - A \cdot \cos{(2\pi \cdot x_i)}
\right),
$
where $A = 10$
\\ \hline

\emph{F9} \newline (Schaffer) &
$-100$ &
$100$  &
$0$    &
$
\func{f}(\vx)
=
\sum_{i=1}^{d-1} (
    0.5 +
    \frac{
        \sin^2{\left( \sqrt{x_i^2 + x_{i+1}^2} \right)} - 0.5
    }{
        \left( 1 + 0.001 (x_i^2 + x_{i+1}^2)\right)^2
    }
)
$
\\ \hline

\emph{F10} \newline (Schwefel) &
$-500$ &
$500$  &
$0.$   &
$
\func{f}(\vx)
=
418.9829 \cdot d - \sum_{i=1}^d x_i \cdot \sin{(\sqrt{|x_i|})}
$
\\ \hline

\end{tabular}
\label{tab:compare_benchmark_list}
\end{table}

\paragraph{Mode size influence.}
To reach high accuracy, we need fine grids.
As we indicated in Section~2.6, in this case, the quantization of the tensor modes seems attractive.
We reshape the original $d$-dimensional tensor $\tj \in \set{R}^{N_1 \times N_2 \times \cdots \times N_d}$ into the tensor $\tilde{\tj} \in \set{R}^{2 \times 2 \times \cdots \times 2}$ of a higher dimension $d \cdot q$, but with smaller modes of size $2$, and apply the \func{TTOpt} algorithm to this ``long'' tensor instead of the original one.

In Table~\ref{tab:benchmark_qtt} we present the comparison of optimization results for the basic algorithm without quantization (``TT'') and for the improved algorithm with quantization (``QTT''). For each value $N$ of the mode size, we choose the number of submodes in the quantized tensor as $q = \log_2 N$.
The QTT-solver gives several orders of magnitude more accurate results than the TT-solver.
At the same time, for the QTT-solver, a regular decrease in the error is observed with an increase in the mode size.
Thus, for the stable operation of gradient-free optimization methods based on the low-rank tensor approximations, it is necessary to quantize the modes of the original tensor.

\paragraph{Rank influence.}
The rank (the size of the maximal-volume submatrices) determines how many points are queried at each iteration of the \func{TTOpt} algorithm, and this parameter is similar to population size in evolutionary algorithms.
Small maximal-volume submatrices may give a better bound for maximal elements (see Eq.~(3) from the main text), but finding small submatrices may be more challenging for the algorithm and may lead to numerical instabilities. 
At the same time, when choosing rank $R$, we should take into account that 
the algorithm will need $ 2 \cdot T \cdot (d q) \cdot P \cdot R^2$ function calls, where $T$ is the number of sweeps (it should be at least $1$, however, for better convergence, it is worth taking values of $4-5$) and $P=2$ is a submode size.
Hence we have inequality $R \leq \sqrt{\frac{M}{4 \cdot T \cdot d \cdot q \cdot}}$, where $M$ is a given limit on the number of function requests.

\begin{figure}[H]
    \vskip 0.2in
    \centering
    \includegraphics[scale=0.32]{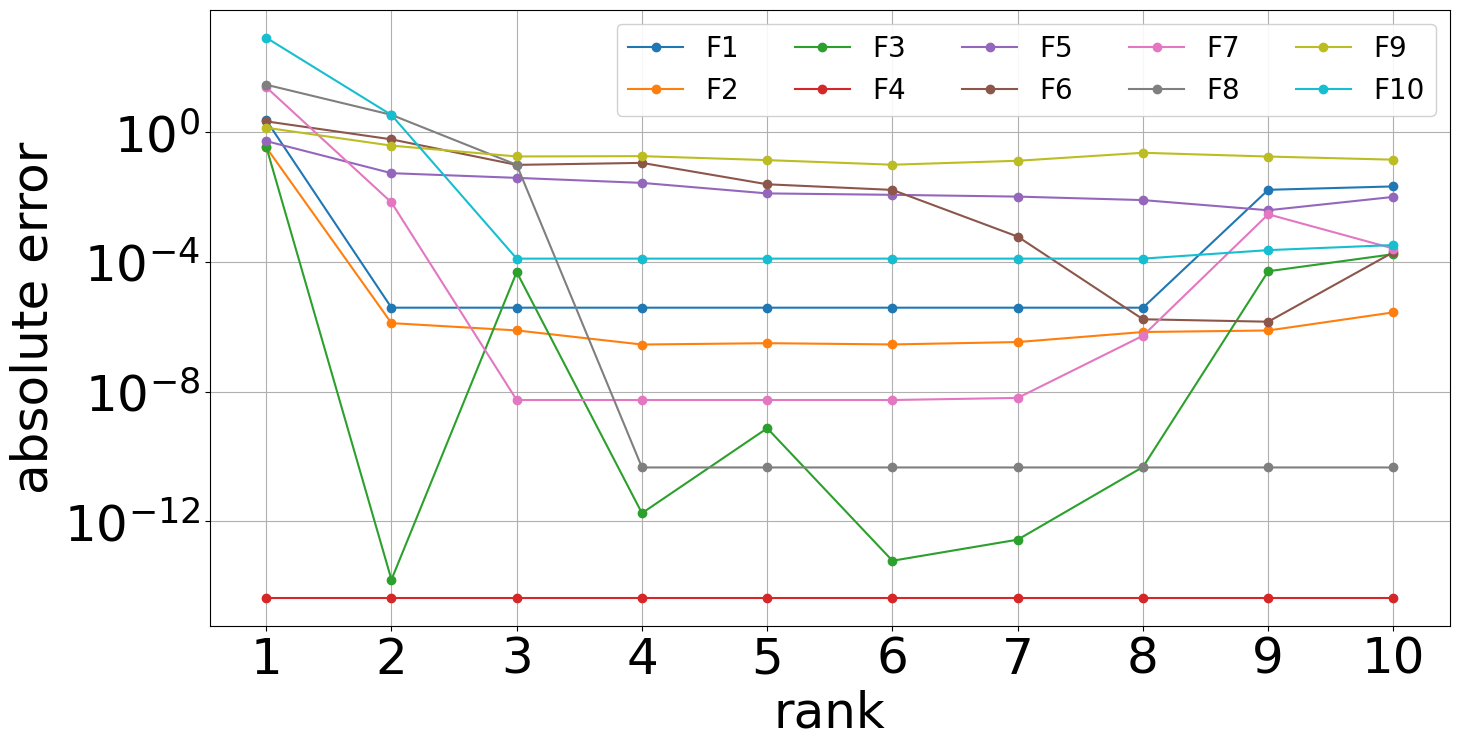}
    \caption{
        The dependence of the final error (absolute deviation of the obtained value from the exact minimum) on the rank for various benchmark functions. 
        The reported values are averaged over ten independent runs.
    }
    \label{fig:benchmark_rank}
    \vskip -0.2in
\end{figure}

In Figure~\ref{fig:benchmark_rank} we demonstrate the dependence of the \func{TTOpt}'s accuracy on the rank.
As can be seen, with small ranks ($1$ or $2$), we have too low accuracy for most benchmarks.
At the same time, the accuracy begins to drop at too high-rank values ($7$ or more), which is due to the insufficient number of sweeps taken by the algorithm for convergence.

\paragraph{Number of function queries influence.}
The number of requests to the objective function can be determined automatically based on algorithm iterations.
Thus, with a total number of sweeps $T$, we will have $\order{T \cdot d \cdot \max_{1 \leq k \leq d}{\left(N_k R_k^2\right)}}$ calls to the objective function.
However, in practice, it turns out to be more convenient to limit the maximum number of function calls, $M$, according to the computational budget.

\begin{figure}[H]
    \vskip 0.2in
    \centering
    \includegraphics[scale=0.32]{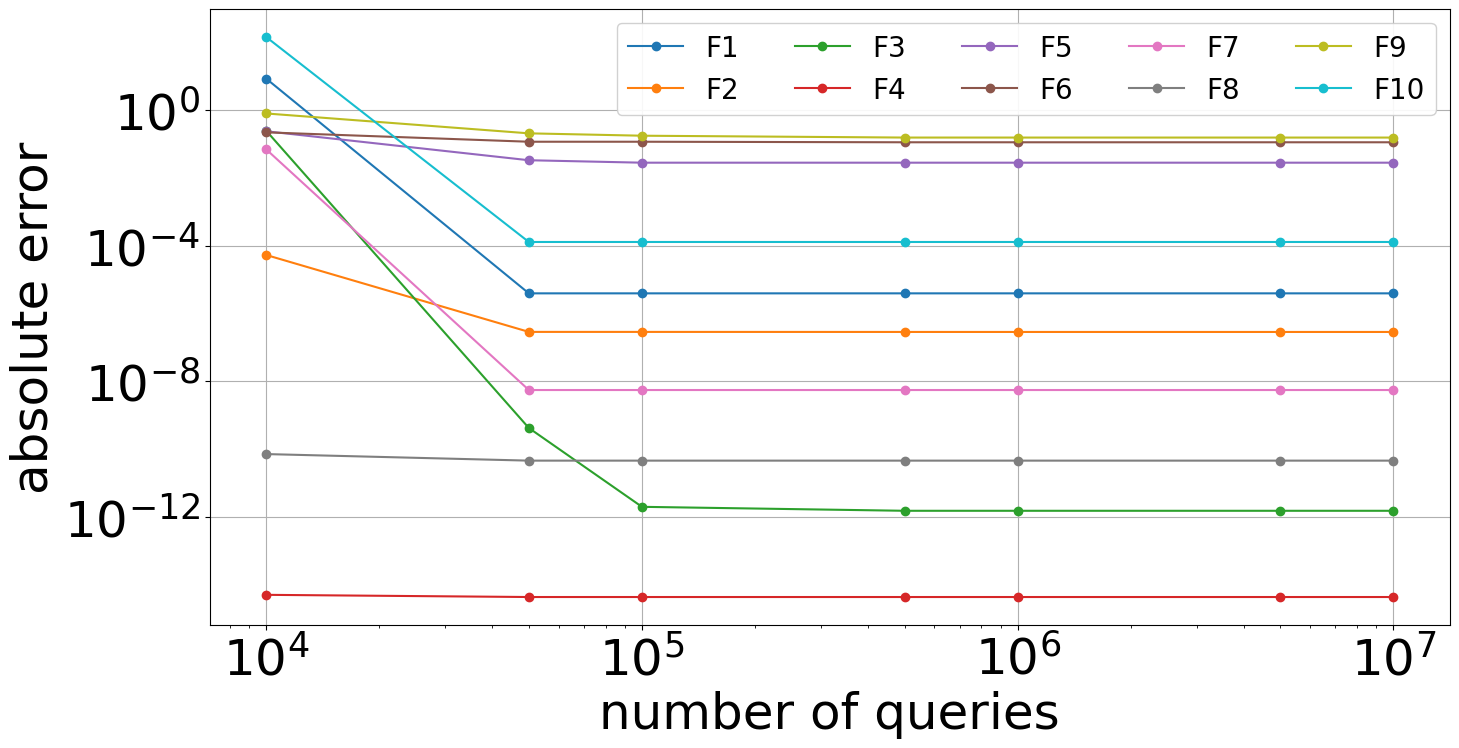}
    \caption{
        The dependence of the final error (absolute deviation of the obtained value from the exact minimum) on the number of target functions calls for various benchmark functions. 
        The reported values are averaged over ten independent runs.
    }
    \label{fig:benchmark_iter}
\end{figure}

In Figure~\ref{fig:benchmark_iter} the dependence of the accuracy on the total number of requests, $M$, to the objective function is presented.
Predictably, as $M$ increases, the accuracy also increases.
The plateau for
benchmarks are associated with the dependence of the result on the remaining parameters ($R$, $q$) of the \func{TTOpt} solver.

\begin{table}
\caption{
    Comparison of the ``direct'' (TT) and ``quantized'' (QTT) \func{TTOpt} solvers in terms of the final error (absolute deviation of the obtained value from the exact minimum) for various benchmark functions. 
    The reported values are averaged over ten independent runs.
}

\begin{center}
\begin{tiny}
\begin{sc}

\begin{tabular}{|p{1.0cm}|p{0.45cm}|p{0.75cm}|p{0.75cm}|p{0.75cm}|p{0.75cm}|p{0.75cm}|p{0.75cm}|p{0.75cm}|p{0.75cm}|p{0.75cm}|p{0.75cm}|}\hline

Mode size
&
& \emph{F1}
& \emph{F2}
& \emph{F3}
& \emph{F4}
& \emph{F5}
& \emph{F6}
& \emph{F7}
& \emph{F8}
& \emph{F9}
& \emph{F10} \\ \hline

\multirow{2}{*}{256}
& TT &  1.2e+00 &  2.0e-02 &  0.0e+00 &  7.7e-05 &  1.0e+00 &  9.8e-02 &  9.4e+01 &  8.0e-01 &  4.2e-01 &  4.5e-01 \\
& QTT &  1.2e+00 &  2.3e-02 &  0.0e+00 &  7.7e-05 &  1.0e+00 &  1.6e-01 &  9.4e+01 &  8.0e-01 &  3.9e-01 &  4.5e-01  \\ \hline 
\multirow{2}{*}{1024}
& TT &  1.6e+01 &  4.2e+00 &  3.0e+01 &  2.6e-01 &  5.7e+01 &  2.0e+00 &  1.9e+10 &  4.7e+01 &  1.5e+00 &  1.1e+03 \\
& QTT &  1.8e-01 &  8.2e-03 &  6.9e-05 &  4.8e-06 &  2.1e-01 &  7.1e-02 &  5.2e+00 &  5.0e-02 &  1.2e-01 &  2.8e-02  \\ \hline 
\multirow{2}{*}{4096}
& TT &  1.9e+01 &  1.5e+01 &  5.0e+08 &  5.1e-01 &  1.4e+02 &  5.7e+00 &  5.1e+10 &  9.8e+01 &  3.5e+00 &  2.6e+03 \\
& QTT &  3.5e-02 &  1.8e-03 &  0.0e+00 &  3.0e-07 &  3.9e-02 &  4.3e-02 &  2.6e-01 &  3.1e-03 &  8.7e-02 &  1.0e-02  \\ \hline 
\multirow{2}{*}{16384}
& TT &  2.0e+01 &  1.9e+01 &  9.9e+17 &  5.9e-01 &  1.7e+02 &  5.9e+00 &  6.3e+10 &  1.2e+02 &  3.8e+00 &  3.2e+03 \\
& QTT &  8.2e-03 &  7.8e-04 &  2.7e-07 &  1.9e-08 &  2.6e-02 &  8.9e-02 &  2.2e-02 &  1.9e-04 &  1.2e-01 &  3.8e-04  \\ \hline 
\multirow{2}{*}{65536}
& TT &  2.0e+01 &  1.9e+01 &  1.2e+10 &  6.7e-01 &  2.2e+02 &  7.8e+00 &  7.1e+10 &  1.6e+02 &  4.4e+00 &  3.6e+03 \\
& QTT &  2.0e-03 &  1.3e-04 &  1.2e-09 &  1.2e-09 &  2.2e-02 &  7.1e-02 &  9.4e-04 &  1.2e-05 &  1.4e-01 &  1.6e-04  \\ \hline 
\multirow{2}{*}{262144}
& TT &  2.1e+01 &  2.3e+01 &  1.7e+16 &  8.0e-01 &  3.0e+02 &  8.4e+00 &  8.9e+10 &  1.8e+02 &  4.5e+00 &  3.7e+03 \\
& QTT &  5.0e-04 &  3.3e-05 &  1.1e-09 &  7.3e-11 &  3.0e-02 &  3.8e-02 &  3.7e-05 &  7.6e-07 &  1.4e-01 &  1.3e-04  \\ \hline 
\multirow{2}{*}{1048576}
& TT &  2.1e+01 &  2.8e+01 &  5.5e+16 &  8.3e-01 &  3.3e+02 &  8.4e+00 &  1.2e+11 &  1.9e+02 &  4.4e+00 &  3.7e+03 \\
& QTT &  1.3e-04 &  1.2e-05 &  0.0e+00 &  4.5e-12 &  1.7e-02 &  6.8e-02 &  5.9e-06 &  4.7e-08 &  1.1e-01 &  1.3e-04  \\ \hline 

\end{tabular}
\end{sc}
\end{tiny}
\end{center}
\vskip -0.1in
\label{tab:benchmark_qtt}
\end{table}

\paragraph{Function dimensionality influence.}
One of the advantages of the proposed approach is the possibility of its application to essentially multidimensional functions.
In Table~\ref{tab:compare_benchmark_dim_s} we present the results of \func{TTOpt} for functions of various dimensions (we removed the \emph{F6} function from benchmarks, since its optima are known only for $2$, $5$ and $10$-dimensional cases).
Note that as a limit on the number of requests of the objective function, we choose $10^4 \cdot d$, and the values of the remaining parameters were chosen the same as above.

As can be seen, even for $500$-dimensional functions, the \func{TTOpt} method results in fairly accurate solutions for most benchmarks. However, for benchmarks \emph{F4}, \emph{F7} and \emph{F9} the errors are larger than for lower dimensions. We suspect that our heuristic of the number of objective function evaluations is not accurate in these cases.

\begin{table}
\caption{The result of the \func{TTOpt} optimizer in terms of the final error $\epsilon$ (absolute deviation of the obtained optimal value relative to the global minimum) and computation time $\tau$ (in seconds) for various benchmark functions and various dimension numbers ($d$).}
\vskip 0.05in

\begin{center}
\begin{small}
\begin{sc}

\begin{tabular}{|p{1.5cm}|p{0.7cm}|p{2.3cm}|p{2.3cm}|p{2.3cm}|p{2.3cm}|}\hline

Function        &
                &
$d \, = \,  10$ &
$d \, = \,  50$ &
$d \, = \, 100$ &
$d \, = \, 500$ \\ \hline

\multirow{2}{*}{\emph{F1}}
& $\epsilon$ 
&  3.9e-06
&  3.9e-06
&  3.9e-06
&  3.9e-06
\\ 
& $\tau$ 
& \textit{3.1}
& \textit{37.8}
& \textit{131.0}
& \textit{3153.5}
\\ \hline 

\multirow{2}{*}{\emph{F2}}
& $\epsilon$ 
&  2.9e-07
&  3.7e-06
&  5.2e-06
&  2.1e-05
\\ 
& $\tau$ 
& \textit{2.5}
& \textit{36.3}
& \textit{129.5}
& \textit{3153.1}
\\ \hline 

\multirow{2}{*}{\emph{F3}}
& $\epsilon$ 
&  2.3e-12
&  4.9e-10
&  1.1e-09
&  4.7e-09
\\ 
& $\tau$ 
& \textit{2.6}
& \textit{36.8}
& \textit{132.6}
& \textit{3205.4}
\\ \hline 

\multirow{2}{*}{\emph{F4}}
& $\epsilon$ 
&  4.4e-15
&  2.2e-14
&  4.4e-14
&  1.0e+00
\\ 
& $\tau$ 
& \textit{2.5}
& \textit{35.6}
& \textit{129.2}
& \textit{3131.1}
\\ \hline 

\multirow{2}{*}{\emph{F5}}
& $\epsilon$ 
&  2.5e-02
&  3.7e-02
&  3.7e-02
&  3.7e-02
\\ 
& $\tau$ 
& \textit{2.5}
& \textit{36.1}
& \textit{130.4}
& \textit{3132.2}
\\ \hline 

\multirow{2}{*}{\emph{F7}}
& $\epsilon$ 
&  5.5e-09
&  8.9e-08
&  3.4e-07
&  5.6e+02
\\ 
& $\tau$ 
& \textit{2.5}
& \textit{35.6}
& \textit{130.2}
& \textit{3123.7}
\\ \hline 

\multirow{2}{*}{\emph{F8}}
& $\epsilon$ 
&  4.6e-11
&  2.3e-10
&  4.6e-10
&  2.3e-09
\\ 
& $\tau$ 
& \textit{2.5}
& \textit{35.6}
& \textit{130.1}
& \textit{3124.0}
\\ \hline 

\multirow{2}{*}{\emph{F9}}
& $\epsilon$ 
&  3.4e-01
&  9.3e-01
&  2.2e+00
&  1.0e+01
\\ 
& $\tau$ 
& \textit{2.5}
& \textit{36.0}
& \textit{130.6}
& \textit{3157.4}
\\ \hline 

\multirow{2}{*}{\emph{F10}}
& $\epsilon$ 
&  1.3e-04
&  6.4e-04
&  1.3e-03
&  6.4e-03
\\ 
& $\tau$ 
& \textit{2.6}
& \textit{35.7}
& \textit{130.0}
& \textit{3140.7}
\\ \hline 


\end{tabular}
\end{sc}
\end{small}
\end{center}
\label{tab:compare_benchmark_dim_s}
\end{table}

\begin{table}
\caption{
    Comparison of the \func{TTOpt} optimizer with Bayesian optimization~\cite{JMLR:v22:18-220} baselines in terms of the final error $\epsilon$ (absolute deviation of the obtained optimal value relative to the global minimum) and computation time $\tau$ (in seconds) for various $10$-dimensional benchmark functions.
    Note that $\tau$ values for Simultaneous Optimistic Optimization (\func{SOO}), Direct Simultaneous Optimistic Optimization (\func{dSOO}),  Locally Oriented Global Optimization  (\func{LOGO}) and Random Optimization (\func{RANDOM}) refers to the time measured for a complied C-code, while our \func{TTOpt} optimizer  is implemented in python, and will be more time-efficient if written in C.}
\vskip 0.15in

\begin{center}
\begin{small}
\begin{sc}

\begin{tabular}{|p{2.15cm}|p{0.45cm}|p{2.2cm}|p{2.2cm}|p{2.2cm}|p{2.2cm}|}\hline

&
& Ackley
& Rastrigin
& Rosenbrock
& Schwefel \\ \hline

\multirow{2}{*}{TTOpt}
& $\epsilon$ &
3.9e-06 & \textbf{4.6e-11} & \textbf{3.9e-01} & \textbf{8.4e-02}
\\
& $\tau$ &
\textit{1.23} & \textit{1.21} & \textit{1.18} & \textit{1.21}
\\ \hline

\multirow{2}{*}{dSOO}
& $\epsilon$ &
\textbf{4.0e-10} & 2.0e+00 & 8.1e+00 & 5.3e+02
\\
& $\tau$ &
\textit{8.10} & \textit{7.20} & \textit{7.75} & \textit{7.01}
\\ \hline

\multirow{2}{*}{SOO}
& $\epsilon$ &
9.0e-10 & 2.29e+00 & 7.0e+02 & 5.2e+02
\\
& $\tau$ &
\textit{7.44} & \textit{7.57} & \textit{0.66} & \textit{7.31}
\\ \hline

\multirow{2}{*}{LOGO}
& $\epsilon$ &
1.2e-09 & 3.44e+01 & 7.9e+00 & 5.3e+02
\\
& $\tau$ &
\textit{6.80} & \textit{0.77} & \textit{7.06} & \textit{7.51}
\\ \hline

\multirow{2}{*}{RANDOM}
& $\epsilon$ &
1.1e+01 & 2.29e+00 & 1.5e+00 & 1.2e+03
\\
& $\tau$ &
\textit{0.78} & \textit{7.34} & \textit{7.25} & \textit{0.77}
\\ \hline

\end{tabular}
\end{sc}
\end{small}
\end{center}
\vskip -0.1in
\label{tab:compare_benchmark_bayesian_10d}
\end{table}

\subsection{Comparison with Bayesian optimization}
\label{sec:appendix_bayesian}

In Table~\ref{tab:compare_benchmark_bayesian_10d} we present the results of \func{TTOpt} and several Bayesian methods for $10$-dimensional benchmarks. We selected functions supported by the Bayesian optimization package from\footnote{
    The source code is available at \url{https://github.com/Eiii/opt_cmp}
}~\cite{JMLR:v22:18-220}.
Note that in all cases we chose $10^5$ as the limit on the number of requests to the objective function and the values of the remaining parameters were chosen the same as above. \func{TTOpt} outperforms all tested Bayesian algorithms for Rastrigin, Rosenbrock, and Schwefel functions. For the Ackley function, the difference in accuracy is not significant. On average, \func{TTOpt} is faster than Bayesian methods, despite they are implemented in \textbf{C} language. We stress that standard Bayesian methods are not applicable in higher-dimensional problems.

\subsection{Formulation of reinforcement learning problem as black-box optimization task}
\label{apx:bbo-rl}

Here we describe a typical reinforcement learning setting within Markov decision process formalism.
The agent acts in the environment that has a set of states $\mathit{S}$. In each state $s \in \mathit{S}$ the agent takes an action from a set of actions $a \in \mathit{A}$. Upon taking this action, the agent receives a local reward $r(s, a)$ and reaches a new state $s^{\prime}$, determined by the transition probability distribution $\mathcal{T}\left(s^{\prime} \mid s, a\right)$. The policy $\pi(a \mid s)$ specifies which action the agent will take depending on its current state. 
Upon taking $T$ ($T$ is also called horizon) actions, the agent receives a cumulative reward, defined as
\begin{equation}
J = \sum_{t=0}^{T-1} \gamma^t r(s_t, a_t),    
\end{equation}
where $\gamma \in [0, 1]$ the is discounting factor, specifies the relevance of historic rewards for the current step. 
\par The goal of the agent is to find the policy $\pi^{\ast}(a \mid s)$ that maximizes the expected cumulative reward $J$ over the agent's lifetime. In policy-based approaches, the policy is approximated by a function $\pi(a \mid s, \vt)$ (for example, a neural network), which depends on a vector of parameters $\vt$. It follows then that the cumulative reward is  a function of the parameters of the agent:
\begin{equation} 
J(\vt) = \mathbb{E}_{(s_t,a_t) \sim  \mathcal{T}, \pi(\vt)} \bigg[\sum_{t=0}^{T-1} \gamma^t r(s_t,a_t))  \bigg],
\end{equation}
where $r(s_t,a_t) \sim 
r(s_{t},\pi (s_{t-1} \mid \vt)$. 
In case of episodic tasks we can assume $\gamma=1$. Finding an optimal policy can be done by maximizing the cumulative reward $J$ with respect to parameters $\vt$:
\begin{equation}
    \pi^{\ast}(a \mid s) = \pi(a \mid s, \vt^{\ast}),
\end{equation}
where $\vt^{\ast} \simeq \text{argmax} ~ J(\vt)$.
Notice that $J$ may be non-differentiable due to the stochastic nature of $\mathcal{T}$ or the definition of $r$, depending on a particular problem formulation. However, this does not pose a problem for direct optimization algorithms.

To summarize, the RL problem can be transformed into a simple optimization problem for the cumulative reward $J(\vt)$. The parameters of this function are the weights of the agent. Optimization of the cumulative reward with direct optimization algorithms is an on-policy learning in RL algorithm classification.

\subsection{Rank dependence study}
\label{apx:rl-ranks}
Since rank is an important parameter of our method, we studied its influence on the rewards in RL, see Figure~5.
Note that the rank determines how many points are queried at each iteration, and this parameter is similar to population size in evolutionary algorithms.
We found almost no dependency of the final reward on rank after $R > 3$ (on average).
The Eq.~(3) from the main text states that small maximal-volume submatrices should give a better bound for the maximal element.
However, finding small submatrices may be more challenging for the algorithm.
It turns out that reward functions in considered RL tasks are "good" for the maximum volume heuristic, e.g., even with small ranks, the algorithm produces high-quality solutions.

\begin{table}[t]
\caption{The mean and standard deviation ( $\mathbb{E} \pm \sigma$) of final cumulative reward before and after fine-tuning with \func{TTOpt}. The policy's weights are from the original repository of ARS ~\cite{NEURIPS2018_ARS}. }
\label{apx:tab:ars}
\begin{center}
\begin{small}
\begin{sc}
\begin{tabular}{|p{3.9cm}|p{4.4cm}| p{4.4cm}|}
\hline
{} & ARS~\cite{NEURIPS2018_ARS} &  ARS  TTOpt($2^8$)  \\
\hline
Ant-v3 &  
4972.48$\pm$21.58 &   
\textbf{5039.90}$\pm$57.00 \\ \hline
HalfCheetah-v3      &  
6527.89$\pm$82.70 &   
\textbf{6840.39}$\pm$87.41 \\ \hline
Hopper-v3 &   
\textbf{3764.74}$\pm$355.08  &   
3296.49$\pm$11.81 \\ \hline
Humanoid-v3  &
11439.79$\pm$51.44  &   
\textbf{11560.01}$\pm$54.08 \\ \hline
Swimmer-v3 &
354.43$\pm$2.32 &   
\textbf{361.87}$\pm$1.76 \\ \hline
Walker2d-v3 &
\textbf{11519.77}$\pm$112.55 &   
11216.25$\pm$88.32 \\ \hline
\end{tabular}
\end{sc}
\end{small}
\end{center}
\end{table}

\begin{table}[t]
\caption{
The number of hidden units in each layer of convolutional policy $h$, the total number of parameters $d$, the sizes of the state and action spaces $A$ and $S$, the rank $R$ and the activation function between the layers (Act.). The average number of function quires per iteration (population size) in the case of TTOpt and ES baselines, respectively, is denoted by $Q$ (the values separated by a comma). The number of seeds is $S_{d}$. 
}
\label{apx:tab:hp}
\begin{center}
\begin{small}
\begin{sc}
    \begin{tabular}{|p{0.3cm}|p{0.3cm}|p{0.3cm}|p{0.3cm}|p{0.3cm}|p{0.3cm}|p{1.0cm}|p{1.0cm}|p{0.8cm}|}
    \hline
    & $H$ &  $D$ & $S$ &  $A$ &  $R$ & $Act.$ & $Q$ & $S_{d}$ \\
    \hline
    S &   8 &  55 &   8 &  2 &  3 &  \func{tanh} & $55 , 64$ & $7$ \\
    \hline
    L &   8 &  55 &   8 &  2 &  3 &  \func{ReLu} & $53,64$ & $7$\\
    \hline
    I &   4 &  26 &   4 &  1 &  3 &  \func{tanh} & $57,64$ & $7$ \\
    \hline
    H &   4 &  44 &  17 &  6 &  5 &  \func{tanh} & $120,128$ & $7$ \\
    \hline
    \end{tabular}
\end{sc}
\end{small}
\end{center}
\end{table}

\subsection{Constraint Handling in Evolutionary Algorithms}
\label{apx:Constraint}
There are two options to satisfy constraints in evolutionary computation called projection and penalization. These steps can be represented as two functions, $\vt_{p} = f_{proj}(\vt)$, and $f_{pen}(\vt_{p},\vt)$ with a regularization term:  
\begin{equation}
    J_p(\vt) = J(\vt_{p}) - \lambda  f_{pen}(\vt_{p},\vt).
\end{equation}
In this work, we use the constraint functions described below.
\emph{\func{CDF} projection} is applied in experiments with mode size $N=3$ (see Table~3 from the main text). The idea is to use the cumulative density function to map normally distributed parameters of the policies to \{-1,0,1\} set:
\begin{equation}
    \vt = 
    \begin{cases}
       -1  & \func{CDF}(\vt) \leq \frac{1}{3}, \\ 
        0  & \frac{1}{3} < \func{CDF}(\vt) < \frac{2}{3}, \\
        1  & \func{CDF}(\vt) \geq \frac{2}{3}. \\ 
    \end{cases}
\end{equation}
\emph{Uniform projection} is applied when $N=256$ in experiments shown in Table~(3) from the main text. In this case, the idea is to keep the value if it satisfies the bounds, otherwise, we draw a new sample uniformly from a grid defined in Algorithm \ref{alg:ttmin}: 
\begin{equation}
    \theta^{i}_{p} = 
    \begin{cases}
        \theta^{i} ,& \text{if} \: L\leq \theta^{i} \leq U,\\
         \vx_i[k] & \text{otherwise}.
    \end{cases}
\end{equation}
\emph{Quadratic penalty} is applied in all experiments. If $L$ and $U$ are the bounds, then $f_{pen}(\vt_{p},\vt) =  \sum_{i:\theta^{i} < L} (L-\theta^{i})^{2} + \sum_{i:\theta^{i} > U} (\theta^{i}-U)^2$. We set $\lambda=0.1$ in all experiments.

\subsection{Reinforcement Learning Experiments} 
\label{apx:main_exp_extra}
  
\begin{figure*}[ht]
    \vskip 0.2in
    \centering
    \includegraphics[scale=0.14]{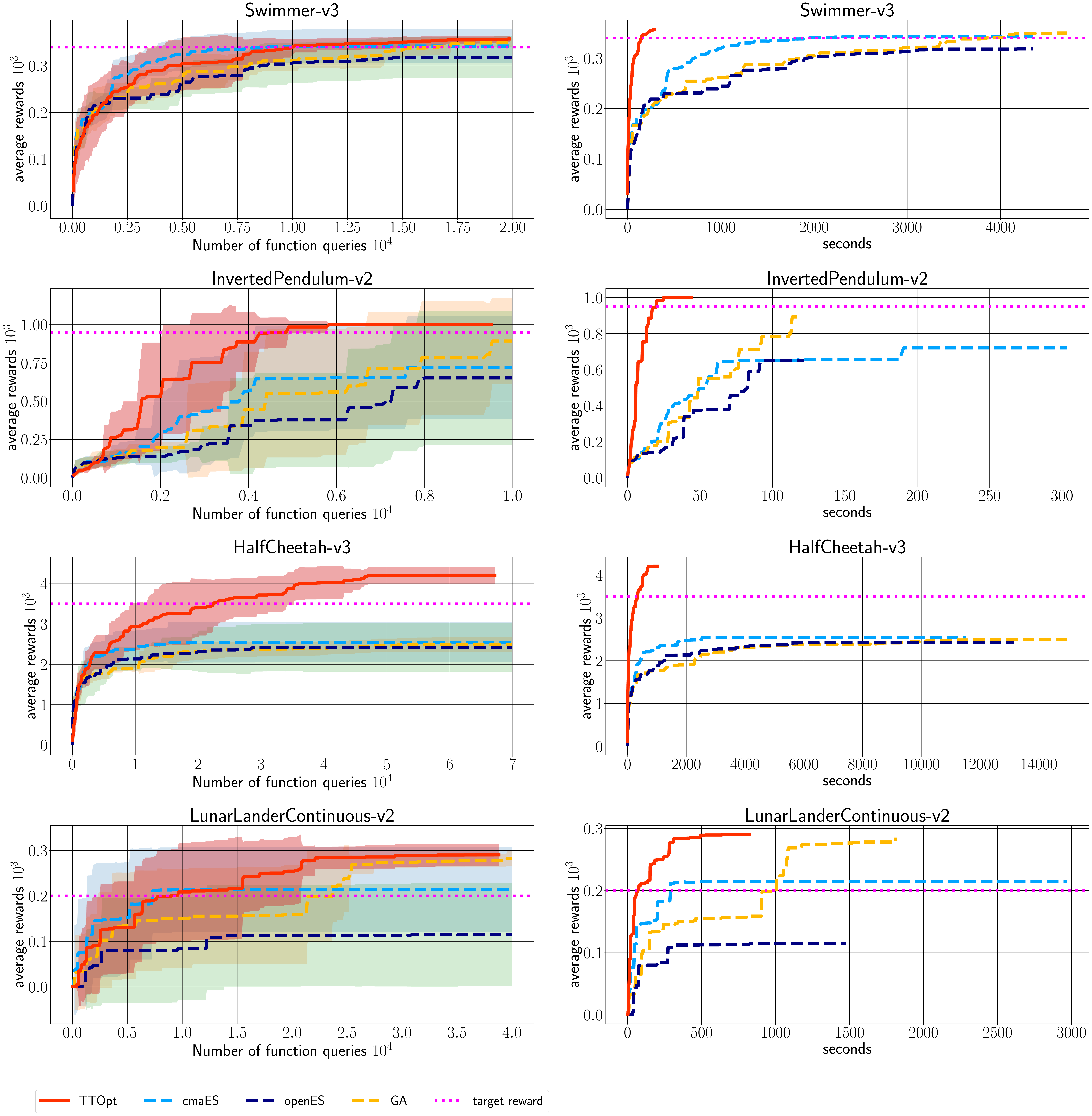}
    \caption{Training curves of \func{TTOpt} and baselines for $N=3$ possible weight values: $(-1, 0, 1)$. Left is the dependence of the average cumulative reward on the number of interactions with the environment (episodes). Right is the same reward depending on the execution time. The reward is averaged for seven seeds. The shaded area shows the difference of one standard deviation around the mean.}
    \label{apx:fig_apx_joint}
    \vskip -0.2in
\end{figure*}

\begin{figure*}[ht]
    \vskip 0.2in
    \centering
    \includegraphics[scale=0.14]{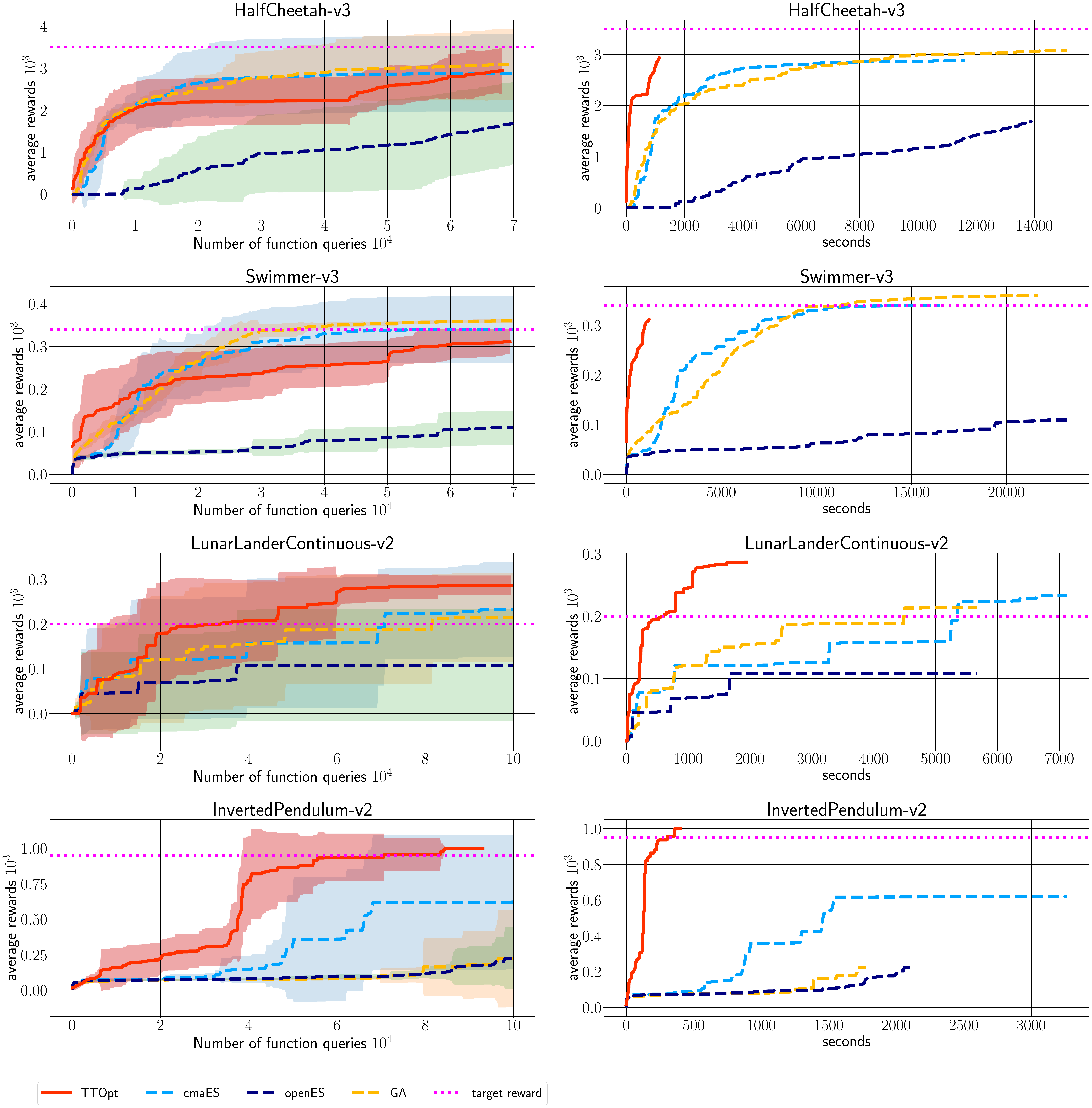}
    \caption{Training curves of \func{TTOpt} and baselines for $N=256$ possible weight values. Left is the dependence of the average cumulative reward on the number of interactions with the environment (episodes). Right is the same reward depending on the execution time. The reward is averaged for seven seeds. The shaded area shows the difference of one standard deviation around the mean.}
    \label{apx:fig_apx_joint256}
    \vskip -0.2in
\end{figure*}

Figure~\ref{apx:fig_apx_joint} and 
Figure~\ref{apx:fig_apx_joint256} 
show training curves for all test environments which were not included in the main text.
\paragraph{Fine-tuning of Linear Policies.}
\label{apx:qtt_and_qtt_ars}
We use \func{TTOpt} to fine-tune Augmented Random Search (ARS)~\cite{NEURIPS2018_ARS} linear policies obtained from the original paper. The cost function is the average of seven independent episodes with fixed random seeds. The upper and lower grid bounds are estimated using statistics of pre-trained linear policies: $b_{i} = \theta_{i} \pm \alpha \cdot \sigma (\vt)$ with $\alpha=0.1$. For Ant, Humanoid ~\cite{ErezTT11}, Walker~\cite{ErezTT11} and HalfCheetah~\cite{4400335} we select $\alpha=0.5$, and for Swimmer~\cite{Coulom-2002a} and Hopper~\cite{doi:10.1177/027836498400300207} we set $\alpha=1$.


\newpage

\newpage

\newpage

\begin{figure*}[ht]
\vskip 0.2in
\centering

\begin{subfigure}
    \centering
    \includegraphics[scale=0.15]{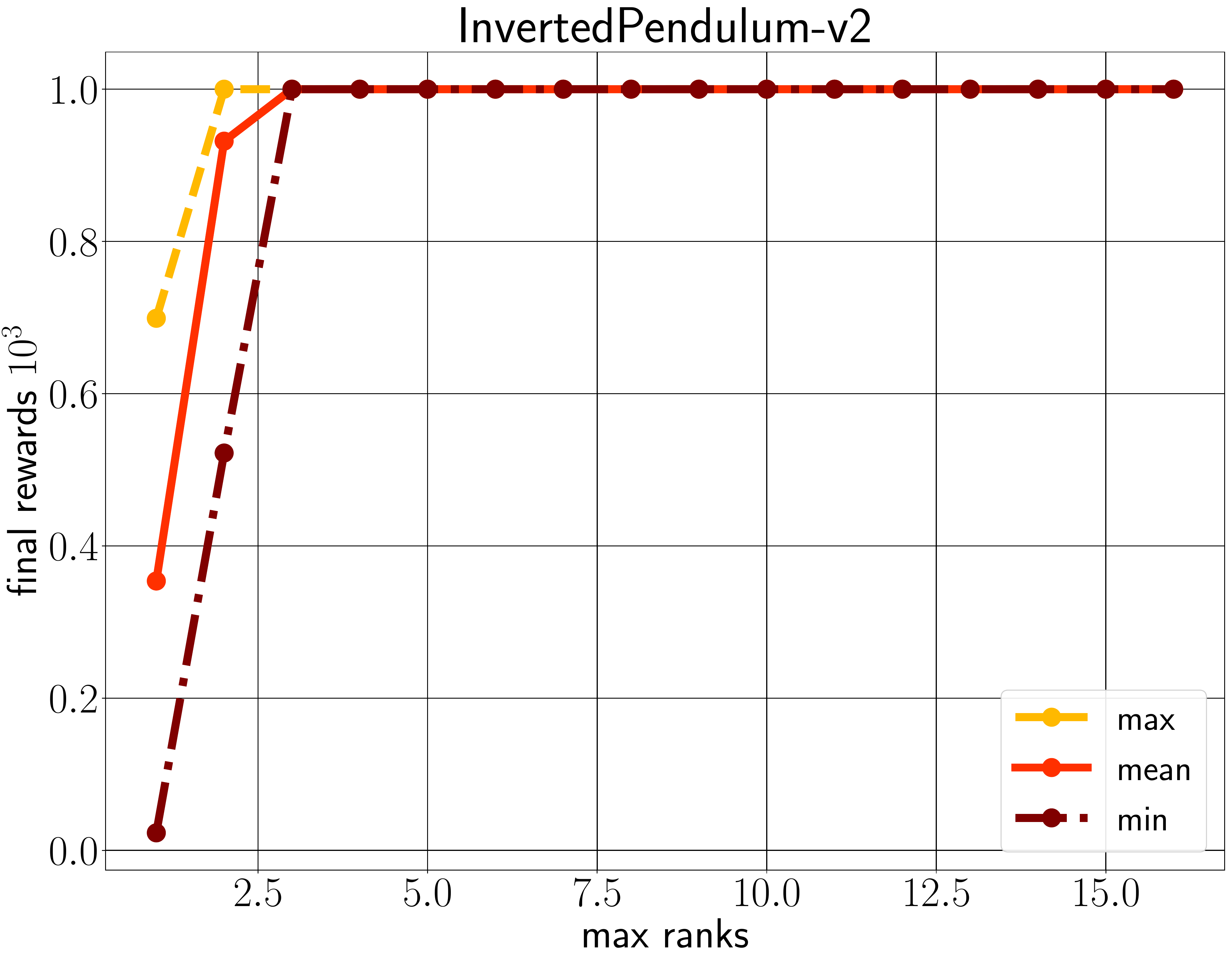}
    \label{fig:sfig_rank_I}
\end{subfigure}
\begin{subfigure}
    \centering
    \includegraphics[scale=0.15]{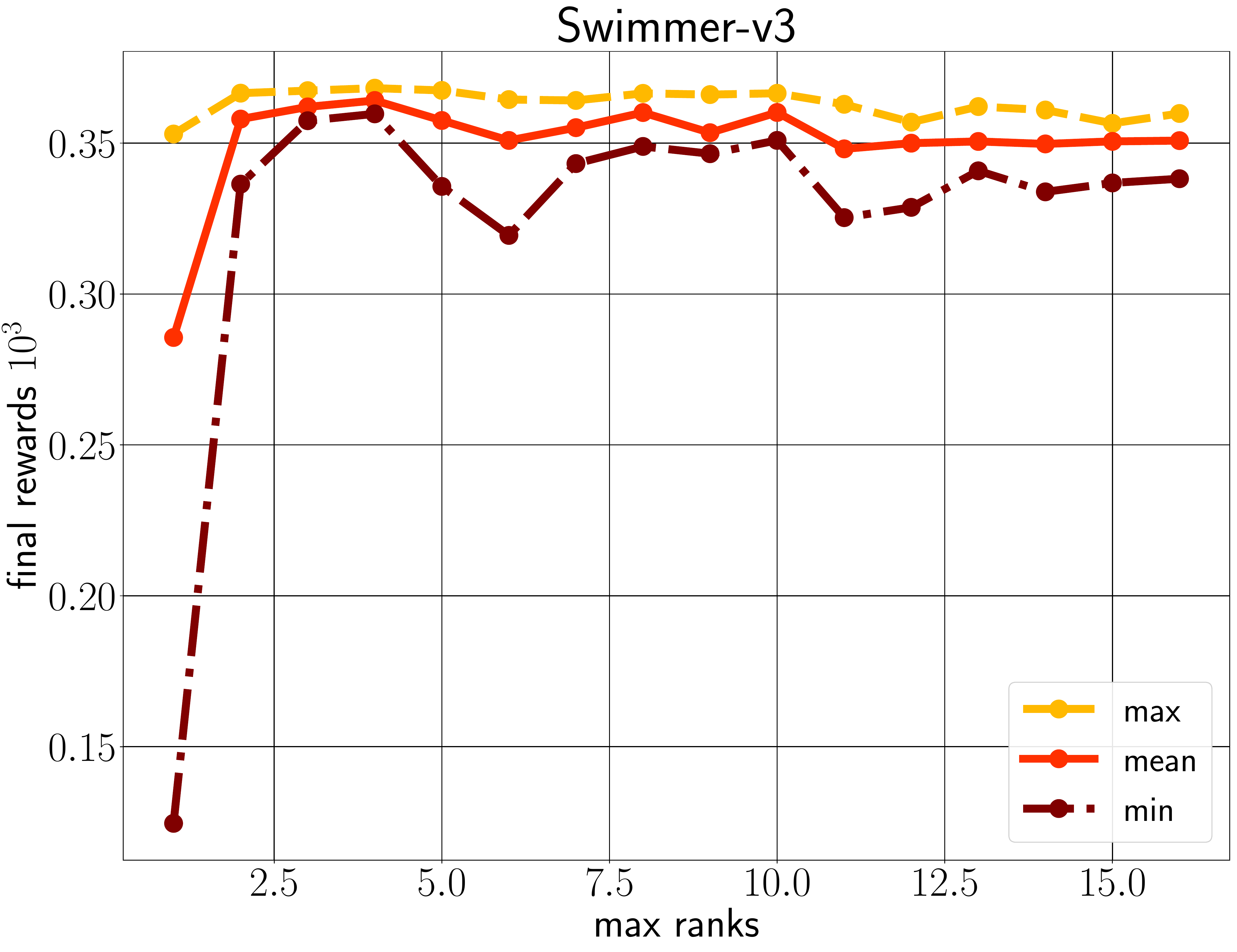}

    \label{fig:sfig_rank_S}
\end{subfigure}

\begin{subfigure}
    \centering
    \includegraphics[scale=0.15]{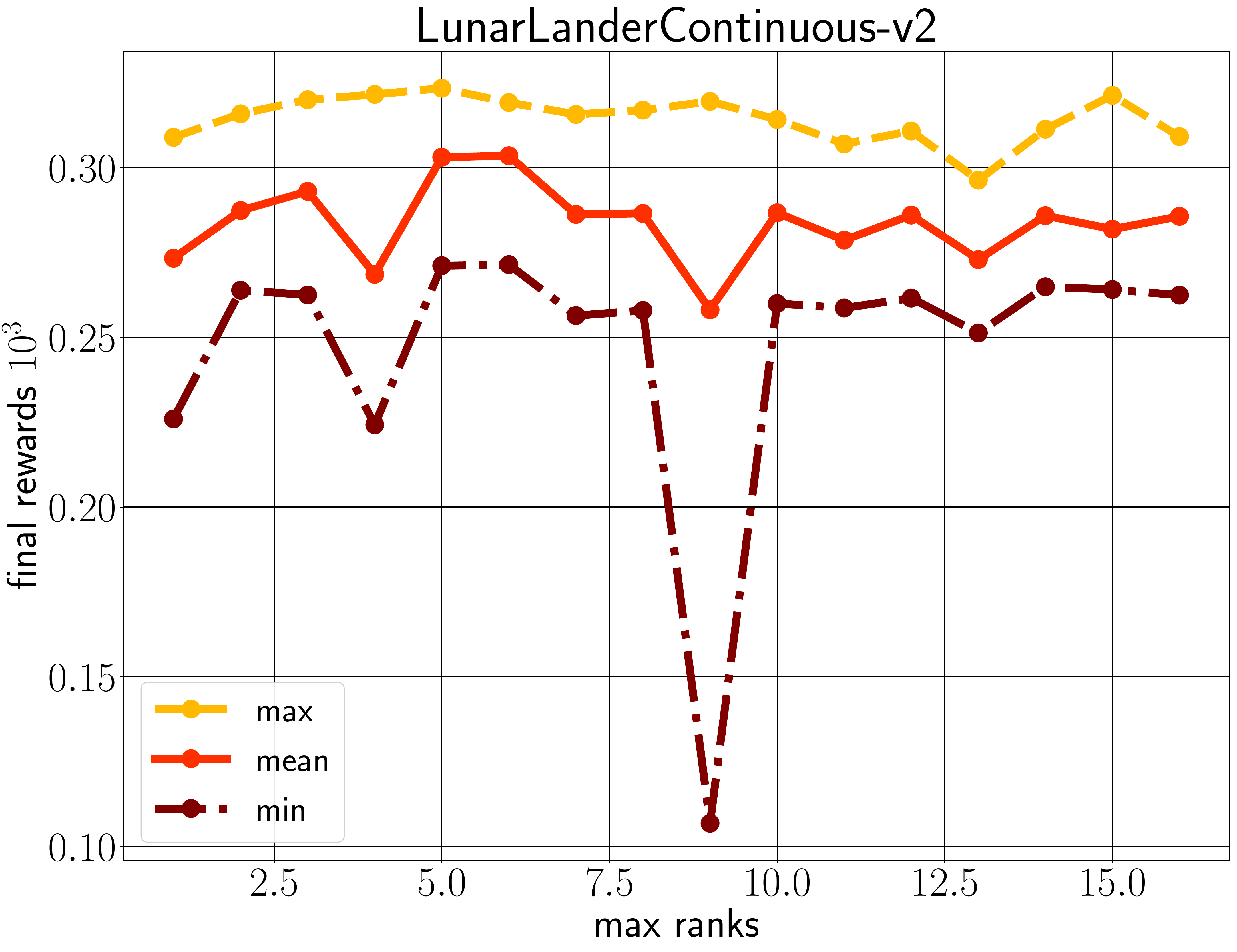}
    \label{fig:sfig_rank_L}
\end{subfigure}
\begin{subfigure}
    \centering
    \includegraphics[scale=0.15]{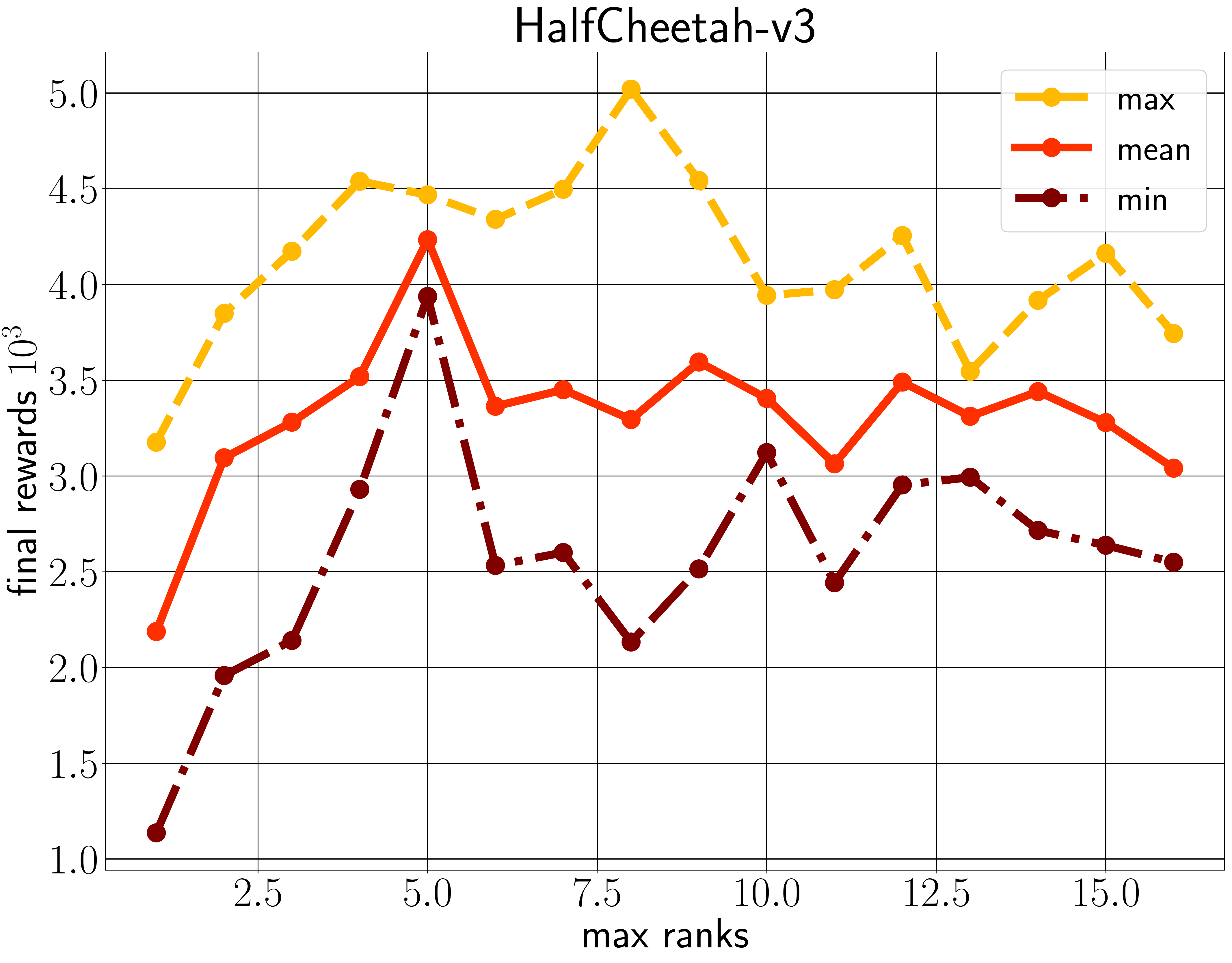}
    \label{fig:sfig_rank_H}
\end{subfigure}
\caption{The dependency of the final cumulative reward on rank. The mean, the minimum, and the maximum over seven random seeds are presented. The mode size is $N=3$.}
\label{apx:fig_ranks}
\vskip -0.2in
\end{figure*}

\end{document}